\documentclass[10pt,twocolumn,letterpaper]{article}

\usepackage{iccv}              % To produce the CAMERA-READY version
\usepackage{colortbl}
\definecolor{iccvblue}{rgb}{0.21,0.49,0.74}
\usepackage[pagebackref,breaklinks,colorlinks,allcolors=iccvblue]{hyperref}

\usepackage{multirow}
\usepackage{adjustbox}
\usepackage{svg}
\usepackage{verbatim}
\def\paperID{5410} % *** Enter the Paper ID here
\def\confName{ICCV}
\def\confYear{2025}

\title{No Pose at All: Self-Supervised Pose-Free 3D Gaussian Splatting \\ from Sparse Views}

\author{Ranran Huang, Krystian Mikolajczyk \\
Imperial College London\\
{\tt\small \{r.huang24, k.mikolajczyk\}@imperial.ac.uk 
}
}
\begin{document}
\maketitle
\begin{abstract}
We introduce SPFSplat, an efficient framework for 3D Gaussian splatting from sparse multi-view images, requiring \textbf{no ground-truth poses} during training or inference. It employs a shared feature extraction backbone, enabling simultaneous prediction of 3D Gaussian primitives and camera poses in a canonical space from unposed inputs within a single feed-forward step. Alongside the rendering loss based on estimated novel-view poses, a reprojection loss is integrated to enforce the learning of pixel-aligned Gaussian primitives for enhanced geometric constraints.
% To reinforce geometric consistency, a reprojection loss is introduced to align Gaussians and the estimated poses of input views.
This pose-free training paradigm and efficient one-step feed-forward design make SPFSplat well-suited for practical applications. Remarkably, despite the absence of pose supervision, SPFSplat achieves state-of-the-art performance in novel view synthesis even under significant viewpoint changes and limited image overlap. It also surpasses recent methods trained with geometry priors in relative pose estimation. 
Code and trained models are available on our project page: \href{https://ranrhuang.github.io/spfsplat/}{https://ranrhuang.github.io/spfsplat/}.
% % \tt\small \url{https://ranrhuang.github.io/spfsplat/}.

\end{abstract}
 \vspace{-10pt}    
\section{Introduction}
\label{sec:intro}

% \begin{figure}[h]
%     \centering
%     \subfloat[Pose-required Methods]{\includegraphics[width=0.5\textwidth]{figs/pipeline_v2_1.pdf}}\hfill
%     \subfloat[Supervised Pose-free Methods]{\includegraphics[width=0.5\textwidth]{figs/pipeline_v2_2.pdf}}\hfill
%     \subfloat[Self-Supervised Pose-free Methods]{\includegraphics[width=0.5\textwidth]{figs/pipeline_v2_3.pdf}}
%     \caption{Comparison}
%     \label{fig:three_figures}
% \end{figure}

Recent advancements in 3D reconstruction and novel view synthesis (NVS) have been driven by Neural Radiance Fields (NeRFs)~\cite{mildenhall2021nerf} and 3D Gaussian splatting (3DGS) ~\cite{kerbl20233dgs}.
A standard NVS training pipeline~\cite{chen2021mvsnerf, xu2024murf, charatan2024pixelsplat, chen2024mvsplat, ye2025noposplat, smart2024splatt3r} reconstructs a 3D scene from input views and optimizes it by aligning rendered novel views with ground-truth images.

\begin{figure}[th]
    \centering
    \includegraphics[width=\linewidth]{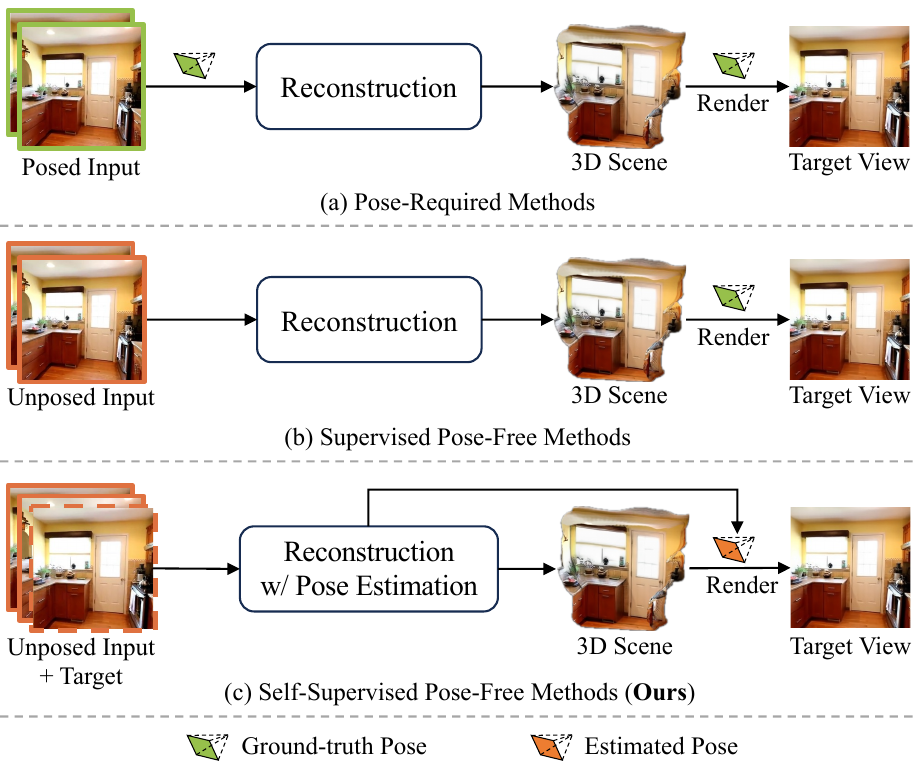}
    \vspace{-15pt}
    \caption{Comparison of three \textbf{training} pipelines for sparse-view 3D scene reconstruction in novel view synthesis. For simplicity, the image rendering loss on the rendered target view is omitted. Our self-supervised pose-free pipeline estimates target-view poses to optimize 3D scene representations reconstructed from unposed images, thereby eliminating the reliance on ground-truth poses during training.}
    \label{fig:pipeline_comparison}
    \vspace{-15pt}
\end{figure}
State-of-the-art (SOTA) methods based on NeRF~\cite{chen2021mvsnerf, xu2024murf} and 3DGS~\cite{chen2024mvsplat, charatan2024pixelsplat, tang2024lgm, zhang2024gslrm, xu2024grm}  typically employ geometry-aware architectures, relying on camera poses estimated using Structure-from-Motion (SfM)~\cite{schonberger2016sfm} to reconstruct 3D scenes, as illustrated in Fig.~\ref{fig:pipeline_comparison} (a). However, the acquisition of camera poses from SfM is computationally expensive and often unreliable in sparse-view scenarios due to insufficient correspondences, limiting the applicability of these \textit{pose-required} methods. To address this, recent research has focused on novel view synthesis under pose-free settings.

\begin{figure*}[ht]
    \centering
    \includegraphics[width=1\textwidth]{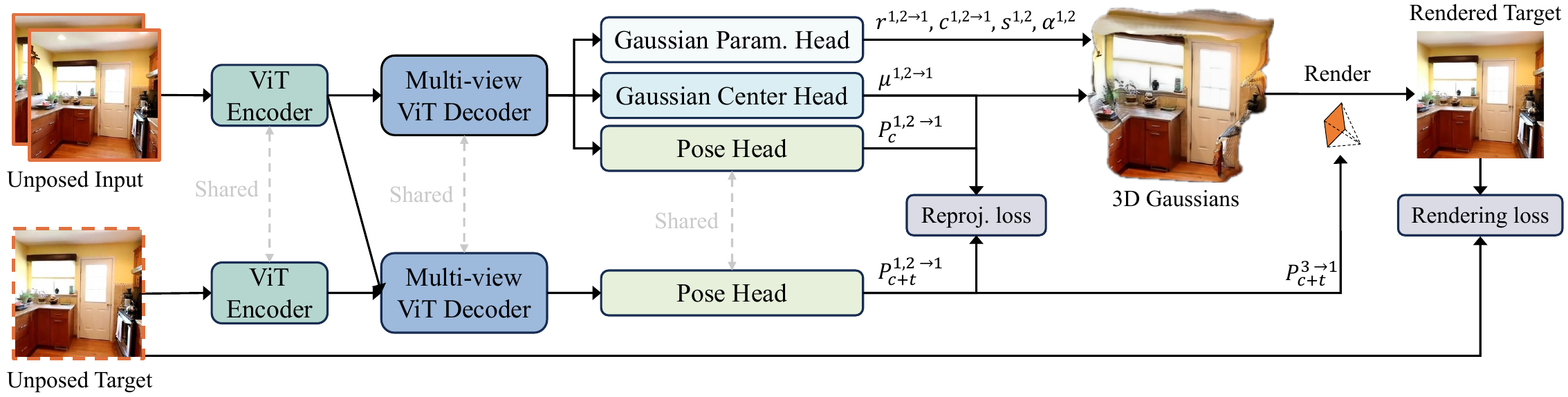}
    \vspace{-15pt}
    \caption{Training pipeline of SPFSplat. Three specialized heads are integrated into a shared ViT backbone, simultaneously predicting Gaussian centers, additional Gaussian parameters, and camera poses from unposed images in a canonical space, where the first input view serves as the reference. Only the context-only branch (above) is used during inference, while the context-with-target branch (below) is employed exclusively during training to estimate target poses, which are used for rendering loss supervision. Additionally, a reprojection loss enforces alignment between Gaussian centers and their corresponding pixels, using the estimated context poses from both branches. Our method jointly optimizes 3D Gaussians and poses, improving geometric consistency and reconstruction quality. }
    \label{fig:overview}
    \vspace{-15pt}
\end{figure*}
Existing pose-free methods reconstruct 3D scenes from unposed images by learning in a canonical space~\cite{jiang2024leap, wang2024pflrm, smart2024splatt3r, ye2025noposplat}, leveraging latent scene representations~\cite{kani2024upfusion, sajjadi2022upsrt}, or jointly optimizing both input-view camera poses and 3D scene representations~\cite{hong2024coponerf, chen2023dbarf, lin2021barf}. Although these methods do not require known input-view poses at inference, they are still trained using rendering losses given ground-truth poses at novel viewpoints, as shown in Fig.~\ref{fig:pipeline_comparison} (b).
We therefore categorize these approaches as \textit{supervised pose-free} methods.
As a result, their training remains confined to datasets with known camera poses, limiting scalability to large-scale real-world data without pose annotations.

This raises a critical question: \textbf{Are ground-truth novel-view poses truly indispensable for optimizing 3D scenes during training}? One solution is to optimize the 3D scenes using poses estimated from the model, referred to as the \textit{self-supervised pose-free} paradigm in Fig.~\ref{fig:pipeline_comparison} (c). However, this presents an inherent challenge: since the rendering loss intrinsically couples the learning of 3D scene geometry and camera poses, pose errors can degrade reconstruction quality, which further hampers pose estimation. Such mutual dependency creates a feedback loop that can potentially lead to unstable training or even divergence. Recent self-supervised pose-free approaches~\cite{hong2024pf3plat, kang2025selfsplat} struggle to mitigate this issue primarily due to their use of separate modules for scene reconstruction and pose estimation, discouraging the learning of consistent feature representations across the two tasks and impairing geometric alignment.
% , making it difficult for pose predictions to leverage global geometric cues from the reconstructed scene, and for the reconstruction to benefit from accurate camera alignment. 
Consequently, these methods still lag far behind SOTA pose-required and supervised pose-free methods~\cite{chen2024mvsplat, charatan2024pixelsplat, ye2025noposplat}.

% Specifically, \cite{hong2024pf3plat, kang2025selfsplat} adopt separate modules for scene reconstruction and pose estimation, which discourages the learning of consistent feature representations and impairs geometric alignment. Moreover, instead of learning pose and depth in an end-to-end manner, \cite{hong2024pf3plat} relies on off-the-shelf initialization, limiting its scalability to large-scale datasets, thereby undermining one of the core advantages of self-supervised learning.

% Their inefficient multi-module sequential pipelines 
% are also highly susceptible to accumulated errors across individual components.
% , leading to degraded 3D reconstruction quality and misaligned estimated poses and 3D Gaussians.
% for 3D reconstruction and pose estimation limit geometric information sharing, causing misalignment between poses and Gaussians. 
% Moreover, pose estimation errors not only lead to inaccurate supervision but also directly degrade 3D reconstruction quality because they adopt the local-to-global paradigm to transform Gaussians from the local camera to the global coordinate frame using the estimated poses at input views.
% This local-to-global paradigm directly introduce pose noises to 3D reconstruction. 
% This issue is exacerbated by the absence of accurate pose supervision during optimization, further limiting the effectiveness of these methods.

% In this work, we propose an efficient self-supervised pose-free training framework for generalizable 3D Gaussian Splatting (3DGS), eliminating the need for ground-truth camera poses during both training and inference.
To address the challenge, we introduce SPFSplat, a \textbf{s}elf-supervised \textbf{p}ose-\textbf{f}ree approach for 3D Gaussian splatting from sparse views, where the novel-view poses are estimated given target images during training. As shown in Fig.~\ref{fig:overview}, SPFSplat employs a shared backbone for feature extraction, equipped with dedicated heads for predicting 3D Gaussian primitives and camera poses. This unified architecture not only enhances computational efficiency but also facilitates joint feature learning for scene reconstruction and pose estimation, improving geometric consistency and mitigating the destabilizing feedback loop. This is achieved by enabling 3D geometry to benefit from accurate, context-aware camera alignment and allowing pose predictions to leverage global scene context, as a form of mutual reinforcement.
Furthermore, we draw inspiration from canonical-space-based methods~\cite{wang2024dust3r, leroy2024mast3r, ye2025noposplat, smart2024splatt3r, jiang2024leap} and directly predict 3D Gaussian primitives relative to the reference view to reduce the impact of pose errors on scene geometry. We also complement the rendering loss with a reprojection loss that explicitly aligns the predicted Gaussians with their corresponding image pixels, imposing stronger geometric constraints to enhance training stability.

We make the following key contributions:
(1) We propose SPFSplat, an efficient framework that simultaneously predicts 3D Gaussians and camera poses in a canonical space from sparse views without requiring any ground-truth pose annotations during training or inference. It incorporates a Gaussian prediction module and a lightweight pose head, enabling joint optimization of scene reconstruction and pose estimation using rendering and reprojection losses.
(2) To the best of our knowledge, SPFSplat is the first self-supervised pose-free method that outperforms SOTA pose-required and supervised pose-free NVS approaches even under extreme viewpoint changes and limited image overlap.
(3) Despite the absence of pose supervision, our feed-forward relative pose estimation is not only highly efficient but also outperforms recent SOTA methods relying on the supervision of geometric priors.

\section{Related Work}
\label{sec:related}
% \begin{figure*}[h]
%     \centering
%     \subfloat[Pose-required Methods]{\includegraphics[width=0.3\textwidth]{figs/framework_comparision1.pdf}}\hfill
%     \subfloat[Supervised Pose-free Methods]{\includegraphics[width=0.3\textwidth]{figs/framework_comparision2.pdf}}\hfill
%     \subfloat[Self-Supervised Pose-free Methods]{\includegraphics[width=0.3\textwidth]{figs/framework_comparision3.pdf}}
%     \caption{Comparison}
%     \label{fig:three_figures}
% \end{figure*}

\subsection{Novel View Synthesis}
NeRF~\cite{mildenhall2021nerf} and 3DGS~\cite{kerbl20233dgs} have shown impressive results in photorealistic NVS. While early methods require dense input views for high-quality output, recent approaches~\cite{chen2021mvsnerf, xu2024murf, charatan2024pixelsplat, chen2024mvsplat, tang2024lgm, zhang2024gslrm, xu2024grm, ye2025noposplat, smart2024splatt3r} focus on 3D reconstruction and novel view synthesis from sparse-view images. The typical NVS pipeline involves reconstructing 3D scenes from input views, followed by optimization with synthesized images aligned to ground-truth targets. Based on their reliance on ground-truth poses during training and inference, existing methods can be categorized into three groups: pose-required, supervised pose-free, and self-supervised pose-free methods, as shown in Fig.~\ref{fig:pipeline_comparison}.

\noindent\textbf{Pose-Required Methods} rely on accurate camera poses during both training and inference to reconstruct 3D scenes using various geometric operations~\cite{chen2021mvsnerf, xu2024murf, charatan2024pixelsplat, chen2024mvsplat, tang2024lgm, zhang2024gslrm, xu2024grm}.
These include constructing cost volumes for multi-view aggregation~\cite{chen2021mvsnerf,xu2024murf}, leveraging epipolar transformers or cost volumes based on feature matching for depth estimation and Gaussian primitive prediction~\cite{charatan2024pixelsplat,chen2024mvsplat}, and encoding camera poses via Pl\"ucker ray embeddings~\cite{zhang2024gslrm, tang2024lgm, xu2024grm}. While effective, such methods rely on Structure-from-Motion (SfM), which is computationally expensive and often unreliable in sparse-view scenarios.
Although recent pose estimation methods~\cite{wang2024dust3r, leroy2024mast3r, zhang2022relpose, lin2024relpose++, wang2023posediffusion} attempt to mitigate these limitations,  they still struggle with low-overlap or texture-less data. Consequently, these pose-required approaches are not applicable in unposed settings.

\noindent\textbf{Supervised Pose-Free Methods} enable 3D reconstruction from unposed images, relaxing the requirement for camera poses at inference time. 
For instance, methods such as~\cite{sajjadi2022upsrt, kani2024upfusion} encode unposed images into latent scene representations, while~\cite{hong2024coponerf, chen2023dbarf, lin2021barf} jointly optimize camera poses and NeRF representations. Approaches like LEAP~\cite{jiang2024leap} and PF-LRM~\cite{wang2024pflrm} define neural volumes in the canonical view’s local camera coordinate system. Similarly, Splatt3R~\cite{smart2024splatt3r} and NoPoSplat~\cite{ye2025noposplat} predict 3D Gaussians in a canonical space. However, these methods still rely on ground-truth camera poses during training through image rendering losses~\cite{jiang2024leap, wang2024pflrm, smart2024splatt3r, ye2025noposplat, sajjadi2022upsrt, kani2024upfusion}, pose prediction loss~\cite{hong2024coponerf}, or coarse initialization~\cite{lin2021barf, truong2023sparf}. Therefore, their training remains limited to datasets with ground-truth poses.

 \noindent\textbf{Self-Supervised Pose-Free Methods} completely eliminate the reliance on ground-truth poses during both training and inference.
For instance, Nope-NeRF~\cite{bian2023nope}, CF-3DGS~\cite{fu2024colmap}, and FlowCam~\cite{smith2023flowcam} reconstruct 3D scenes and estimate camera poses incrementally by re-rendering dense video sequences. However, they are limited to continuous video frames and do not generalize well to sparse views.
% InstantSplat~\cite{fan2024instantsplat} initializes Gaussian means and camera poses using a frozen DUSt3R~\cite{wang2024dust3r} backbone and jointly optimizes Gaussian attributes and camera parameters. However, its performance is constrained by the frozen feature extractor.
Recent self-supervised pose-free methods, such as PF3plat~\cite{hong2024pf3plat} and SelfSplat~\cite{kang2025selfsplat} attempt to estimate camera poses from sparse views. Specifically, PF3plat relies on off-the-shelf feature descriptors (LightGlue~\cite{lindenberger2023lightglue}) and RANSAC-based pose initialization, resulting in a pipeline that is neither efficient nor end-to-end trainable. In contrast, 
SelfSplat employs cross-view U-Nets~\cite{ronneberger2015unet} to predict depth and pose separately. Despite their differences, both methods adopt separate modules for pose estimation and scene reconstruction, leading to unshared and inconsistent features, poor geometric alignment, and increased computational overhead. Inaccurate poses further directly degrade reconstruction by corrupting the lifted 3D points, amplifying reconstruction errors and exacerbating the feedback loop instability.

% These methods adopt a local-to-global paradigm: they first predict per-pixel depth, lift Gaussian parameters to local camera coordinates, and then transform them into world coordinates using the estimated poses. 
% The reconstructed Gaussians are optimized via an image rendering loss, combined with either off-the-shelf feature descriptor supervision~\cite{hong2024pf3plat} or RGB and depth consistency constraints~\cite{kang2025selfsplat}.
% However, their local-to-global paradigm and sequential multi-module pipeline lead to error propagation, misalignment between Gaussians and poses, and inefficient training and inference.

Our method also adopts a self-supervised, pose-free paradigm, requiring no ground-truth poses during training or inference. In contrast to~\cite{hong2024pf3plat, kang2025selfsplat}, we jointly optimize 3D Gaussians and camera poses via a shared backbone in a canonical space~\cite{wang2024dust3r, leroy2024mast3r, smart2024splatt3r, ye2025noposplat}, guided by both image rendering and reprojection losses.  This unified design ensures that pose estimation is informed by the same scene geometry that drives Gaussian prediction, encouraging geometric alignment between the scene representation and the predicted poses and enhancing training stability.

\subsection{Structure-from-Motion (SfM)}
Structure-from-Motion (SfM) is a fundamental problem in computer vision~\cite{hartley2003multiple}, involving the simultaneous reconstruction of sparse 3D maps and estimation of camera parameters from a set of images. 
%A traditional SfM pipeline typically begins with keypoint matching across multiple images~\cite{bay2008speeded, lowe2004distinctive, rublee2011orb} to establish geometric relationships using two-view epipolar geometry or homography, often incorporating RANSAC for robust estimation. This is followed by triangulation to derive 3D points. New images are incrementally registered by solving the Perspective-n-Point (PnP) problem, followed by further triangulation and bundle adjustment to jointly refine 3D coordinates and camera parameters.
Recent advances have integrated learning-based approaches into various SfM components. Enhancements include more robust feature descriptors~\cite{detone2018superpoint, Dusmanu2019d2,huang2023drkf}, improved image matching~\cite{sarlin2020superglue}, detector-free matching~\cite{sun2021loftr}, and neural bundle adjustment~\cite{lin2021barf}. Moreover, fully differentiable SfM pipelines have been introduced~\cite{wang2024vggsfm, wang2024dust3r}. Unlike VGGSfM~\cite{wang2024vggsfm}, which focuses on end-to-end sparse reconstruction, DUSt3R~\cite{wang2024dust3r} enables dense 3D reconstruction without requiring camera parameters and has been successfully applied to pose estimation, monocular depth estimation and 3D reconstruction. As an extension, MASt3R~\cite{leroy2024mast3r} integrates feature matching with DUSt3R and improves local feature representation.

Similar to SfM methods, our approach jointly predicts 3D points and camera poses. The image rendering and reprojection losses used during training can be interpreted as a form of bundle adjustment, further jointly refining and aligning the estimated scene representation and poses.

\section{Method}
\label{sec:method}

\subsection{Problem Formulation}
% We aim to learn a feed-forward network $f_{\boldsymbol{\theta}}$ to reconstruct 3D Gaussians from $\textit{N}$ unposed input images $\{\boldsymbol{I}^v\}_{v=1}^{N}$ and synthesize photorealistic images $\boldsymbol{I}^t$ from novel view points.
We aim to learn a feed-forward network that reconstructs 3D Gaussians from $\textit{N}$ unposed input images $\{\boldsymbol{I}^v\}_{v=1}^{N}$ while simultaneously estimating the camera poses. During training, the 3D Gaussians are optimized by synthesizing photorealistic images $\hat{\boldsymbol{I}^t}$ from the estimated poses at target view $t$, thereby eliminating the need for ground-truth poses.

\noindent\textbf{3D Gaussian Reconstruction.}
Following~\cite{smart2024splatt3r, ye2025noposplat}, we predict 3D Gaussians in a canonical 3D space where the first input view $\boldsymbol{I}^1$ serves as the global reference coordinate frame. The network is formulated as:
\begin{equation}
    f_{\boldsymbol{\theta}} : \{\boldsymbol{I}^v\}_{v=1}^{N} \mapsto \{   \boldsymbol{\mathcal{G}}^{v \rightarrow 1}\}_{\substack{v=1,\dots,N}}, 
\label{eq:gaussian_formulation}
\end{equation}
where  $\boldsymbol{\mathcal{G}}^{v \rightarrow 1} = \{(\boldsymbol{\mu}_j^{v \rightarrow 1},  \boldsymbol{r}_j^{v \rightarrow 1},  \boldsymbol{c}_j^{v \rightarrow 1}, \alpha_j^v, \boldsymbol{s}_j^v)\}_{\substack{j=1,\dots,H \times W}}$ represents the pixel-aligned Gaussians for $\boldsymbol{I}^v$, represented in the coordinate frame of $\boldsymbol{I}^1$. 
% and $ v \rightarrow 1 $ 
% indicates that the Gaussian parameters in view $v$ are in the coordinate frame of $\boldsymbol{I}^1$. 
Each Gaussian is parameterized by center $ \boldsymbol{\mu} \in \mathbb{R}^3 $, rotation quaternion $\boldsymbol{r} \in \mathbb{R}^4 $, scale $\boldsymbol{s} \in \mathbb{R}^3 $, opacity $ \alpha \in \mathbb{R} $, and spherical harmonics (SH) $ \boldsymbol{c} \in \mathbb{R}^k $ with \( k \) degrees of freedom. 

% Once the 3D Gaussians are reconstructed, images from novel viewpoints are rendered and then compared against the ground-truth RGB image $\boldsymbol{I}^t$ to compute the image rendering loss, which is used to optimize the Gaussians. Although the canonical space allows different views to be aligned without explicit camera poses, prior works~\cite{smart2024splatt3r, ye2024noposplat} still require ground-truth poses to synthesize images during training.
\noindent\textbf{Pose Estimation.}
% While the canonical space allows different views to be aligned without explicit camera poses, 
% previous methods \cite{smart2024splatt3r, ye2024noposplat} still require ground-truth poses for image synthesis for optimizing Gaussians. To build a truly pose-free training framework, 
We introduce a pose network $f_{\boldsymbol{\phi}}$ to estimate the relative transformation from the view $\boldsymbol{I}^v$ to the reference view $\boldsymbol{I}^1$. 
The estimated relative transformation from $\boldsymbol{I}^v$  to $\boldsymbol{I}^1$ is denoted as $\boldsymbol{P}^{v \to 1} =[\boldsymbol{R}^{v \to 1} | \boldsymbol{T}^{v \to 1} ]$, where $\boldsymbol{R}^{v \to 1} \in \mathbb{R}^{3 \times 3}$ represents the rotation matrix, and $\boldsymbol{T}^{v \to 1} \in \mathbb{R}^{3 \times 1}$ represents the translation vector.  The pose estimation is formulated as:
\begin{equation}
    \boldsymbol{P}^{v \to 1} = f_{\boldsymbol{\phi}}(\boldsymbol{I}^{v}, \boldsymbol{I}^{1}).
\label{eq:pose_head}
\end{equation}

\noindent\textbf{Novel View Synthesis.} During training, to eliminate the reliance on ground-truth poses for image synthesis at target viewpoints, we also estimate the relative pose from $\boldsymbol{I}^t$  to $\boldsymbol{I}^1$, which is denoted as $\boldsymbol{P}^{t \to 1} =[\boldsymbol{R}^{t \to 1} | \boldsymbol{T}^{t \to 1} ]$. Using the estimated transformation, we render images from novel viewpoints following Eq. \ref{eq:image_rendering}:
\begin{equation}
    \begin{aligned}
        \hat{\boldsymbol{I}^t} &= \mathcal{R}(\boldsymbol{P}^{t \to 1}, \{\boldsymbol{\mathcal{G}}^{v \rightarrow 1}\}_{\substack{v=1,\dots,N}}).
    \end{aligned}
\label{eq:image_rendering}
\end{equation}
% Our method is flexible and can be trained with or without camera intrinsic parameters. 
As in prior work, we assume that intrinsic parameters are available from camera sensor metadata. 

\subsection{Architecture}
As shown in Fig.~\ref{fig:overview}, our method consists of four main components: an encoder, a decoder, a pose head, and Gaussian prediction heads. Both the encoder and decoder are based on Vision Transformer (ViT) architectures~\cite{dosovitskiy2021vit}.

\noindent\textbf{Encoder and Decoder.}
For each input view, the RGB image is first patchified and flattened into a sequence of image tokens. Following~\cite{ye2025noposplat}, to alleviate scale ambiguity, we encode the intrinsic camera parameters into an additional token using a linear layer, which is then concatenated with the image tokens along the spatial dimension. It is notable that this operation is optional. As demonstrated in Sec.~\ref{sec:experiments_ablation}, our method surpasses existing approaches even without injecting intrinsic parameters into the backbone.
Each view’s tokens are first processed individually by a weight-sharing ViT encoder. Next, a ViT decoder equipped with cross-attention aggregates multi-view information. 
The decoder jointly reasons over token representations across all input views, with each view attending to all other views, facilitating cross-view information exchange to capture spatial relationships and the global 3D scene geometry. In contrast to pairwise architectures~\cite{wang2024dust3r, leroy2024mast3r}, this approach supports efficient incorporation of additional input views without significantly increasing the memory or computation cost.

\noindent\textbf{Gaussian Prediction Heads.}
Following~\cite{smart2024splatt3r, ye2025noposplat}, we use two DPT-based heads~\cite{ranftl2021dpt} to infer Gaussian parameters.
The first head processes decoder tokens from context views and predicts 3D coordinates for each pixel, defining Gaussian centers. The second head estimates rotation, scale, opacity, and SH coefficients for each Gaussian primitive. As proposed in~\cite{charatan2024pixelsplat, chen2024mvsplat, ye2025noposplat}, we incorporate high-resolution skip connections by feeding the original input image into the prediction heads, preserving fine-grained spatial details.

\noindent\textbf{Pose Head.}
The pose head enables the prediction of poses for input views in a single feed-forward step, and is essential for self-supervised learning of Gaussians. It is built on the same decoder as the Gaussian heads, encouraging shared geometric knowledge and better alignment between the Gaussians and poses. 

Within the pose head, token representations from the encoder and decoder are concatenated, unpatchified, and processed via global average pooling to generate a compact geometric embedding for each view. This embedding is then fed into a lightweight 3-layer MLP, which directly outputs the camera pose as a 10-dimensional representation~\cite{chen2024map}.
The pose is decomposed into translation and rotation for each view. The translation is represented using four homogeneous coordinates~\cite{brachmann2023accelerated}, while the rotation is encoded in a 6D format, capturing two unnormalized coordinate axes. These axes are normalized and combined via a cross-product operation to construct a full rotation matrix~\cite{zhou2019continuity}.
To compute the relative pose with respect to the reference view, the 10D pose representation is converted into a homogeneous transformation matrix $\boldsymbol{P}^{v \to 1} \in \mathbb{R}^{4 \times 4}$. We normalize the camera poses by assigning the first input view the canonical pose $[\mathbf{U} | \mathbf{0}]$, where $\mathbf{U}$ represents the identity matrix, and $\mathbf{0}$ denotes the zero translation vector. More details can be found in the appendix.

During training, to enable image synthesis at target views without ground-truth poses, we introduce an additional input branch that incorporates both context and target views. The encoder tokens from these views are jointly aggregated by the multi-view ViT decoder, after which the target poses are predicted via the pose head.
Importantly, Gaussian reconstruction and target pose prediction are decoupled to prevent information leakage. As illustrated in Fig.~\ref{fig:overview}, Gaussian representations are predicted solely from the context views, while target pose estimation leverages information from both context and target views to achieve a more comprehensive understanding of the global geometric structure. This design ensures that the information from the target view does not influence the 3D Gaussian representation, thereby improving generalization to novel viewpoints.

\subsection{Loss Function}
% transform each pixel to world coordinates using camera intrinsics, predicted depth and accurate camera poses
\noindent\textbf{Image Rendering Loss.}
Our model is trained using ground-truth target RGB images as supervision. The training loss is formulated as a weighted combination of the $L_2$ loss and the LPIPS loss~\cite{zhang2018lpips}, formulated as:
\begin{equation}
    \mathcal{L}_{\text{render}} = \| \boldsymbol{I}^t - \hat{\boldsymbol{I}^t}\|_2 + \gamma \text{LPIPS}(\boldsymbol{I}^t, \hat{\boldsymbol{I}^t}),
\label{eq:rendering_loss}
\end{equation}
where $\boldsymbol{I}^t$ and $\hat{\boldsymbol{I}^t}$ denote the ground-truth and rendered target images, respectively, and $\gamma$ is a weighting factor that balances perceptual similarity and pixel-level accuracy.

\noindent\textbf{Reprojection Loss.}
Existing approaches enforce pixel-aligned Gaussian prediction by constraining Gaussian locations along the input viewing rays~\cite{charatan2024pixelsplat, chen2024mvsplat, xu2024grm, zhang2024gslrm, hong2024pf3plat, kang2025selfsplat}. On the other hand, canonical-space-based methods~\cite{smart2024splatt3r, ye2025noposplat} rely on ground-truth camera poses to guide the canonical 3D points (Gaussian centers).
Both strategies ensure alignment between each pixel and its corresponding 3D point. However, since our model learns 3D Gaussian centers in a canonical space without known camera poses, the network lacks explicit geometric constraints to enforce pixel-aligned Gaussian representation.

A naive solution is to include context views in the rendering loss (Eq. \ref{eq:rendering_loss}) by synthesizing images from them and computing the loss against their ground-truth counterparts. However, this leads to unstable training due to overfitting. Specifically, the network prioritizes improving the rendering quality of the first context view, as the 3D Gaussian space is defined in its camera coordinate, making its rendering independent of the learnable poses. 
Since the Gaussians from this view already capture sufficient scene information, the model suppresses the contribution of other context views by shifting their Gaussian centers away and adjusting camera poses, ultimately leading to training collapse.

To address this issue, we employ a pixel-wise reprojection loss to jointly optimize 3D points and camera poses~\cite{brachmann2024scr, schonberger2016sfm}. Unlike image-based supervision, reprojection loss enforces explicit geometric constraints, reducing overfitting to context views. Specifically, for each pixel $\mathbf{p}^v_j$ in context view $v$, we project the corresponding 3D Gaussian center $\boldsymbol{\mu}_{j}^{v \rightarrow 1}$ from the first camera coordinate frame into 2D pixel coordinates using the estimated pose of view $v$ and minimize the pixel-wise reprojection error. Since context poses can be obtained from both the context-only and context-with-target branches during training (Fig. \ref{fig:overview}), we enforce consistency by applying reprojection loss to both:
% \vspace{-13pt}
\begin{align}
\mathcal{L}_{\text{reproj}} &= \sum_{v=1}^{N} \sum_{j=1}^{H\times W} 
\!\!\left\| \mathbf{p}^v_{j} - \pi(\boldsymbol{K}^v, \boldsymbol{P}^{v \rightarrow 1} , \boldsymbol{\mu}_{j}^{v \rightarrow 1}) \right\|,
\end{align}
% \begin{align}
% \mathcal{L}{\text{reproj}} &= \sum{v=1}^N \sum_{j=1}^{H\times W} \left| \mathbf{p}^v_{j} - \pi(\boldsymbol{K}^v, \boldsymbol{T}^{v \rightarrow 1}, \boldsymbol{\mu}{j}^{v \rightarrow 1}) \right| \
% & + \sum{v=1}^N \sum_{j=1}^{H\times W} \left| \mathbf{p}^v_{j} - \pi(\boldsymbol{K}^v, \boldsymbol{T}c^{v \rightarrow 1}, \boldsymbol{\mu}{j}^{v \rightarrow 1}) \right|
% \end{align}
where  ${\boldsymbol{P}^{v \rightarrow 1} \in \{ \boldsymbol{P}_{c}^{v \rightarrow 1}, \boldsymbol{P}_{c+t}^{v \rightarrow 1}\}}$, $\pi$ represents the camera projection function, $\boldsymbol{K}^v$ denotes the camera intrinsics of view $v$. $\boldsymbol{P}_c^{v \rightarrow 1}$ is the relative pose from view $v$ to the canonical frame estimated from the context-only branch, while $\boldsymbol{P}_{c+t}^{v \rightarrow 1}$ is estimated from both context and target views.
By leveraging the reprojection loss, our method enables stable training and efficient optimization of pixel-aligned 3D Gaussians, without requiring ground-truth camera poses.
\section{Experiments}
\label{sec:experiments}

\begin{table*}[!ht]
    \footnotesize
    \centering
    \resizebox{\textwidth}{!}{ % Resize to fit within page width
    \begin{tabular}{lccccccccccccc}
    \toprule
    \multirow{2}{*}{\textbf{Method}} & \multicolumn{3}{c}{\textbf{Small}} & \multicolumn{3}{c}{\textbf{Medium}} & \multicolumn{3}{c}{\textbf{Large}} & \multicolumn{3}{c}{\textbf{Average}} & \textbf{Time}\\
    \cmidrule(lr){2-4} \cmidrule(lr){5-7} \cmidrule(lr){8-10} \cmidrule(lr){11-13} \cmidrule(lr){14-14}& PSNR $\uparrow$ & SSIM $\uparrow$ & LPIPS $\downarrow$ 
    & PSNR $\uparrow$ & SSIM $\uparrow$ & LPIPS $\downarrow$ 
    & PSNR $\uparrow$ & SSIM $\uparrow$ & LPIPS $\downarrow$ 
    & PSNR $\uparrow$ & SSIM $\uparrow$ & LPIPS $\downarrow$ & (s)\\ 
    \midrule
    \rowcolor{gray!20} \multicolumn{14}{l}{\textit{Pose-Required}} \\ 
    pixelSplat & 20.277 & 0.719 & 0.265 
    & 23.726 & 0.811 & 0.180 
    & 27.152 & 0.880 & 0.121 
    & 23.859 & 0.808 & 0.184 & 0.152 \\
    MVSplat & 20.371 & 0.725 & 0.250  
    & 23.808 & 0.814 & 0.172 
    & 27.466 & 0.885 & 0.115
    & 24.012 & 0.812 & 0.175 & 0.059 \\
    \midrule
    \rowcolor{gray!20} \multicolumn{14}{l}{\textit{Supervised Pose-Free}} \\ 
    % DUSt3R & 14.101 & 0.432 & 0.468 & 15.419 & 0.451 & 0.432 & 16.427 & 0.453 & 0.402 & 15.382 & 0.447 & 0.432 \\
    % MASt3R & 16.305 & 0.516 & 0.451 
    % & 18.106 & 0.561 & 0.377 
    % & 17.975 & 0.524 & 0.402
    % & 17.617 & 0.539 & 0.403\\
    CoPoNeRF & 17.393 & 0.585 & 0.462 
    & 18.813 & 0.616 & 0.392 
    & 20.464 & 0.652 & 0.318 
    & 18.938 & 0.619 & 0.388 & -\\
    Splatt3R & 17.789 & 0.582 & 0.375 
    & 18.828 & 0.607 & 0.330 
    & 19.243 & 0.593 & 0.317 
    & 18.688 & 0.337 & 0.596 & 0.042 \\
    % NopoSplat & 21.085 & 0.721 & 0.237 
    % & 23.132 & 0.776 & 0.185 
    % &	25.082	& 0.818 & 0.141	
    % & 23.187 & 0.774 & 0.185  \\
    NoPoSplat$^\ast$ & 22.514 & 0.784 & 0.210 
    & 24.899 & 0.839 & 0.160 
    & 27.411 & 0.883 & 0.119 
    & 25.033 & 0.838 & 0.160 & 0.042\\
    \midrule
    \rowcolor{gray!20} \multicolumn{14}{l}{\textit{Self-Supervised Pose-Free}} \\ 
    SelfSplat & 14.828 & 0.543 & 0.469 
    & 18.857 & 0.679 & 0.328 
    & 23.338 & 0.798 & 0.208
    & 19.152 & 0.680 & 0.328 & 0.101\\
    PF3plat & 18.358 & 0.668 & 0.298
    & 20.953 & 0.741 & 0.231 
    & 23.491 & 0.795 & 0.179 
    & 21.042 & 0.739 & 0.233 & 1.171 \\
    
    \textbf{SPFSplat} & \underline{22.897}	& \underline{0.792}	& \underline{0.201}	& \underline{25.334} & \underline{0.847} & \underline{0.153}
    & \underline{27.947} & \underline{0.894} & \textbf{0.110}
    & \underline{25.484} & \underline{0.847} & \underline{0.153} & 0.044\\
    \textbf{SPFSplat$^\ast$ }  & \textbf{23.178} & \textbf{0.796} & \textbf{0.200}	
    & \textbf{25.695} & \textbf{0.853} & \textbf{0.151}
    & \textbf{28.377} & \textbf{0.899} & \underline{0.111}	
    & \textbf{25.845} & \textbf{0.852} & \textbf{0.152} & 0.044 \\

    \bottomrule
    \end{tabular}
    }
    \vspace{-5pt}
    \caption{Performance comparison of novel view synthesis on the RE10K dataset~\cite{zhou2018stereo}. The reported runtime reflects only the time required to reconstruct 3D Gaussians from two input images. Our method achieves computational efficiency comparable to NoPoSplat while significantly outperforming previous state-of-the-art pose-required and pose-free methods across all overlap settings, especially in low-overlap scenes. The \textbf{best} and \underline{second-best} results are highlighted. $\ast$ indicates the use of evaluation-time pose alignment (EPA) strategy.}
    \label{tab:rek_results}
    \vspace{-5pt}
\end{table*}
\begin{table*}[!ht]
\footnotesize

\centering

\resizebox{\textwidth}{!}{ % Resize to fit within page width

\begin{tabular}{lcccccccccccc}
\toprule
\multirow{2}{*}{\textbf{Method}} & \multicolumn{3}{c}{\textbf{Small}} & \multicolumn{3}{c}{\textbf{Medium}} & \multicolumn{3}{c}{\textbf{Large}} & \multicolumn{3}{c}{\textbf{Average}} \\
\cmidrule(lr){2-4} \cmidrule(lr){5-7} \cmidrule(lr){8-10} \cmidrule(lr){11-13}& PSNR $\uparrow$ & SSIM $\uparrow$ & LPIPS $\downarrow$ 
& PSNR $\uparrow$ & SSIM $\uparrow$ & LPIPS $\downarrow$ 
& PSNR $\uparrow$ & SSIM $\uparrow$ & LPIPS $\downarrow$ 
& PSNR $\uparrow$ & SSIM $\uparrow$ & LPIPS $\downarrow$ \\ 
\midrule
\rowcolor{gray!20} \multicolumn{13}{l}{\textit{Pose-Required}} \\ 
pixelSplat & 22.088 & 0.655 & 0.284
& 25.525 & \underline{0.777} & 0.197
& 28.527 & 0.854 & 0.139 
& 25.889 & 0.780 & 0.194 \\
MVSplat & 21.412 & 0.640 & 0.290  
& 25.150 & 0.772 & 0.198
& 28.457 & 0.854 & 0.137
& 25.561 & 0.775 & 0.195 \\
\midrule
% DUSt3R & 14.101 & 0.432 & 0.468 & 15.419 & 0.451 & 0.432 & 16.427 & 0.453 & 0.402 & 15.382 & 0.447 & 0.432 \\
% MASt3R & 18.230 & 0.502 & 0.448
% & 19.501 & 0.525 & 0.415 
% & 19.265 & 0.509 & 0.440 &
% 19.186 & 0.516 & 0.429 \\
\rowcolor{gray!20} \multicolumn{13}{l}{\textit{Supervised Pose-Free}} \\ 
CoPoNeRF & 18.651 & 0.551 & 0.485
& 20.654 & 0.595 & 0.418 
& 22.654 & 0.652 & 0.343 
& 20.950 & 0.606 & 0.406 \\
Splatt3R & 17.419 & 0.501 & 0.434 
& 18.257 & 0.514 & 0.405 
& 18.134 & 0.508 & 0.395 
& 18.060 & 0.510 & 0.407 \\
% NopoSplat & 21.200 & 0.609 & 0.308 
% & 23.700 & 0.700 & 0.231 
% &	25.576	& 0.755 & 0.179	
% & 23.863 & 0.702 & 0.228  \\
NoPoSplat$^\ast$ & \underline{23.087} & \underline{0.685} & \underline{0.258} 
& \underline{25.624} & \underline{0.777} & 0.193 
& 28.043 & 0.841 & 0.144 
& 25.961 & \underline{0.781} & 0.189 \\
\midrule
\rowcolor{gray!20} \multicolumn{13}{l}{\textit{Self-Supervised Pose-Free}} \\ 
SelfSplat & 18.301 & 0.568 & 0.408
& 21.375 & 0.676 & 0.314 
& 25.219 & 0.792 & 0.214
& 22.089 & 0.694 & 0.298 \\
PF3plat & 18.112 & 0.537 & 0.376 
& 20.732 & 0.615 & 0.307 
& 23.607 & 0.710 & 0.228
& 21.206 & 0.632 & 0.293 \\

\textbf{SPFSplat} & 22.667 & 0.665 & 0.262
& 25.620 & 0.773 & \underline{0.192}
& \underline{28.607} & \underline{0.856} & \underline{0.136} 
& \underline{26.070} & \underline{0.781} & \underline{0.186} \\
\textbf{SPFSplat$^\ast$ }  & \textbf{23.676} &	\textbf{0.708} & \textbf{0.243}	
& \textbf{26.351} & \textbf{0.801} & \textbf{0.182}	
& \textbf{29.170} & \textbf{0.870} & \textbf{0.131}	
& \textbf{26.796} & \textbf{0.807} & \textbf{0.176} \\
\bottomrule
\end{tabular}
}
\vspace{-5pt}
\caption{Performance comparison of novel view synthesis on the ACID dataset~\cite{liu2021infinite}. The \textbf{best} and \underline{second best} results are highlighted.}
\vspace{-15pt}
\label{tab:acid_results}
\end{table*}
We report evaluation results for quality of novel view synthesis and cross-dataset generalization, as well as pose estimation on several datasets.
% \begin{figure*}[htbp]
%     \centering
%     \includegraphics[width=0.76\textwidth]{figs/cross-dataset.pdf}
%     \caption{Cross-dataset. Our model can better zero-shot transfer to out-of-distribution data than SOTA methods.}
%     \label{fig:cross_dataset}
% \end{figure*}
\subsection{Experimental Settings}
\label{sec:experimental_settings}
\textbf{Dataset.} We train and evaluate our method on the RealEstate10K (RE10K)~\cite{zhou2018stereo} which contains large-scale real estate videos from YouTube, and ACID~\cite{liu2021infinite} dataset which features nature scenes captured by aerial drones. Camera poses for both datasets are obtained via SfM, and we follow the official train-test split used in prior works~\cite{chen2024mvsplat, charatan2024pixelsplat, ye2025noposplat}.
Following~\cite{ye2025noposplat}, we evaluate our method under varying camera overlaps, categorizing input pairs based on image overlap ratios: small (0.05\%–0.3\%), medium (0.3\%–0.55\%), and large (0.55\%–0.8\%), determined using a pretrained dense image matching method~\cite{edstedt2024roma}. To study the impact of training data size, we incorporate the DL3DV dataset~\cite{ling2024dl3dv}, an outdoor dataset with 10K videos and diverse camera motions beyond RE10K.
For cross-dataset generalization, we follow~\cite{chen2024mvsplat, ye2025noposplat} and evaluate on the object-centric DTU dataset~\cite{jensen2014dtu}.

\noindent\textbf{Baselines.}
For novel view synthesis, we compare to baselines including pose-required methods (pixelSplat~\cite{charatan2024pixelsplat} and MVSplat~\cite{chen2024mvsplat}), supervised pose-free methods (CoPoNeRF~\cite{hong2024coponerf}, Splatt3R~\cite{smart2024splatt3r}, and NoPoSplat~\cite{ye2025noposplat}) and self-supervised methods (SelfSplat~\cite{kang2025selfsplat} and PF3plat~\cite{hong2024pf3plat}). For pose estimation, we compare to Superpoint~\cite{detone2018superpoint} + SuperGlue \cite{sarlin2020superglue}, DUSt3R~\cite{wang2024dust3r}, MASt3R~\cite{leroy2024mast3r}, and splatting-based methods including NoPoSplat, SelfSplat and PF3plat.

\noindent\textbf{Evaluation Protocol.} We evaluate novel view synthesis with the standard metrics: pixel-level PSNR, patch-level SSIM~\cite{wang2004ssim}, and feature-level LPIPS~\cite{zhang2018lpips}. As in previous works \cite{sarlin2020superglue, ye2025noposplat}, for pose estimation, we report the area under the cumulative pose error curve (AUC) at thresholds of 5$^\circ$, 10$^\circ$, 20$^\circ$, where the pose error is the maximum of the angular errors
in rotation and translation.

During evaluation for NVS, the target images are typically rendered using ground-truth poses~\cite{chen2024mvsplat, charatan2024pixelsplat,hong2024coponerf, smart2024splatt3r}. A different strategy is to render novel views using the estimated target poses, as in PF3plat~\cite{hong2024pf3plat} and SelfSplat~\cite{kang2025selfsplat}. 
Alternatively, NoPoSplat~\cite{ye2025noposplat} employs an evaluation-time pose alignment (EPA) strategy, which optimizes the target pose during evaluation while keeping the reconstructed Gaussians frozen, such that the rendered image closely matches the ground truth.
Unless otherwise specified, we use estimated poses for a comprehensive evaluation and also report EPA results for a fair comparison with NoPoSplat.
EPA decouples rendering quality from pose estimation, allowing direct assessment of Gaussian reconstruction. In contrast, rendering with estimated poses jointly evaluates both reconstruction quality and the alignment between the estimated poses and the learned Gaussians.
% For rendering target views during evaluation, there are three different strategies for target poses: ground-truth poses, optimal poses and estimated poses.
% Pose-required methods~\cite{chen2024mvsplat, charatan2024pixelsplat} and some supervised pose-free methods~\cite{hong2024coponerf, smart2024splatt3r} utilize ground-truth target poses for evaluation.
% NoPoSplat~\cite{ye2025noposplat} adopts evaluation-time pose alignment (EPA) strategy, optimizing the target camera pose during evaluation. Specifically, after reconstructing 3D Gaussians, the Gaussians are frozen, and only the target pose is optimized to ensure that the rendered image most closely matches the ground truth.
% Self-supervised methods PF3plat~\cite{hong2024pf3plat} and SelfSplat~\cite{kang2025selfsplat} render new views using the predicted target poses during evaluation.

% By utilizing EPA strategy, rendering target images with the optimal poses decouple rendering quality from pose estimation, allowing direct evaluation of Gaussian reconstruction quality. In contrast, when rendering target images using estimated poses, the rendering quality depends on the quality of Gaussian reconstruction as well as the alignment of the estimated poses with the reconstructed Gaussians.
% Unless otherwise specified, we use estimated poses for a more compound evaluation of our method. To ensure a fair comparison with NoPoSplat, we also report results using EPA strategy.

\subsection{Implementation Details.}
Our method is implemented in PyTorch, using a CUDA-based 3DGS renderer with gradient support for camera poses. All models are trained on a single A100 GPU. The encoder follows a ViT-Large architecture with a patch size of 16, and the decoder is ViT-Base. The encoder, decoder, and Gaussian center head are initialized with pretrained MASt3R~\cite{leroy2024mast3r} weights. The pose head is initialized to approximate the identity rotation matrix for stable convergence. All remaining layers are randomly initialized. The loss weights for LPIPS and reprojection loss are set to 0.05 and 0.001, respectively. All experiments are conducted at 256$\times$256 resolution.

\subsection{Results}
\begin{figure*}[ht]
    \centering
    \includegraphics[width=1.0\textwidth]{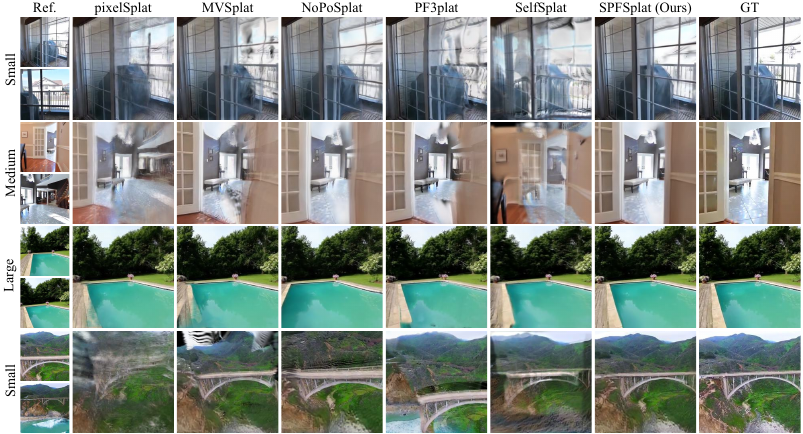}

    \vspace{-5pt}
    \caption{Qualitative comparison on RE10K (top three rows) and ACID (bottom row). Compared to baselines, our method 1) reduces misaligned blending artifacts and ghosting effects, 2) better handles extreme viewpoint changes and texture-less areas (e.g. window), and 3) preserves overall scene geometry (e.g. bridge) and finer details (e.g. swimming pool).}
    \label{fig:qualitative_comparison}
    \vspace{-5pt}
\end{figure*}

\textbf{Novel View Synthesis.} We present quantitative results in Tab.~\ref{tab:rek_results} and Tab.~\ref{tab:acid_results}. Our model outperforms all SOTA methods, including pose-required and supervised pose-free approaches. Notably, it achieves superior results even in cases of small input image overlap and extreme viewpoint changes, as illustrated in Fig.~\ref{fig:qualitative_comparison}, despite the fact that no ground-truth poses were used during training. Furthermore, even without evaluation-time pose alignment (EPA), our model still surpasses NoPoSplat, which indicates that our estimated poses are well aligned with the Gaussians. We report a reconstruction time of 0.044 seconds for 3D Gaussians from two 256$\times$256 input images, making it approximately 3.5$\times$ and 27$\times$ faster than pixelSplat and PF3plat, respectively, on the same A6000 GPU. 
SPFSplat achieves high efficiency by directly constructing Gaussians in a canonical space using a feed-forward network. In contrast, PF3plat relies on separate modules and computationally expensive local feature matching for pose estimation to lift depth predictions into 3D Gaussians.

\begin{figure*}[ht]
    \centering
    \subfloat[Cross-Dataset Generalization: RE10K → ACID]{\includegraphics[width=0.5\textwidth]{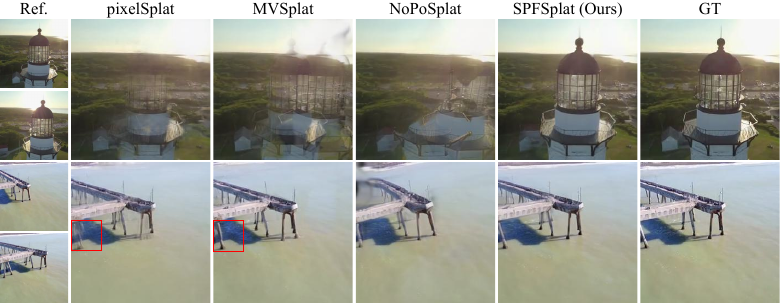}}\hfill
    \subfloat[Cross-Dataset Generalization: RE10K → DTU]{\includegraphics[width=0.5\textwidth]{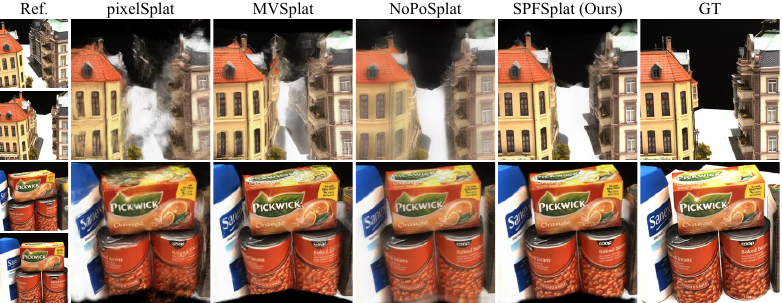}}\hfill
    \vspace{-7pt}
    \caption{Cross-dataset generalization. Some failure regions are highlighted by red rectangles for visual reference. }
    \label{fig:cross_dataset}
    \vspace{-15pt}
\end{figure*}

\noindent\textbf{Relative Pose Estimation.} 
We evaluate pose estimation between two input  images on RE10K and ACID, as shown in Tab.~\ref{tab:pose_estimation}. The evaluation details of baselines are provided in the appendix.
% To assess generalization, the evaluation on ACID is conducted by using the model trained only on RE10K for all splat-based methods. 
Our method supports two pose estimation strategies: direct regression via the pose head, and estimation via PnP~\cite{hartley2003multiple} with RANSAC~\cite{fischler1981ransac}, using the predicted 3D Gaussian centers. Both yield similarly strong results, indicating accurate alignment between the estimated poses and the reconstructed 3D points.
Notably, despite being trained without geometry priors, our method significantly outperforms recent approaches, including MASt3R, from which our model is initialized, demonstrating that our framework effectively optimizes both camera poses and 3D structure using only image-level supervision. 
\begin{table}[!ht]
\footnotesize
\centering
\setlength{\tabcolsep}{5.2pt}

\begin{tabular}{lccccccccccc}
    \toprule
    \multirow{2}{*}{\textbf{Method}} & \multicolumn{3}{c}{\textbf{RE10K}} & \multicolumn{3}{c}{\textbf{ACID}}  \\
    \cmidrule(lr){2-4} \cmidrule(lr){5-7}  
    
    &  5$^\circ$ $\uparrow$ & 10$^\circ$ $\uparrow$ & 20$^\circ$ $\uparrow$ &  5$^\circ$ $\uparrow$ &  10$^\circ$ $\uparrow$ &  20$^\circ$ $\uparrow$  \\
    \midrule
    SP + SG & 0.234 & 0.406 & 0.569 & 0.228 & 0.363 & 0.500 \\
    DUSt3R & 0.336 & 0.541 & 0.702 & 0.118 & 0.279 & 0.470 \\ 
    MASt3R & 0.281	& 0.494 & 0.671	& 0.138 &	0.312 & 0.507 \\
    % \midrule
    NoPoSplat & 0.572 & 0.728 & \underline{0.833}  & 0.335 & 0.497 & 0.645 \\
    SelfSplat  & 0.207 & 0.392 & 0.576 & 0.205 & 0.363 & 0.531 \\
    PF3plat  & 0.187 & 0.398 & 0.613 & 0.060	& 0.165	& 0.340 \\

    \textbf{SPFSplat (PnP)} & \underline{0.613} & \underline{0.754} &	\textbf{0.845} & \underline{0.355} & \underline{0.516} & \underline{0.658} \\
    \textbf{SPFSplat} & \textbf{0.617} & \textbf{0.755}	& \textbf{0.845} & \textbf{0.364} & \textbf{0.520}	& \textbf{0.662} \\
    \bottomrule
    \end{tabular}
    \vspace{-7pt}
    \caption{Pose estimation performance in AUC with various thresholds on RE10K and ACID datasets. To assess generalizability, we evaluate on ACID using the models trained only on RE10K for all splat-based methods. Our method achieves the best results in both in-domain and out-of-domain settings.}
    \label{tab:pose_estimation}
    \vspace{-15pt}
\end{table}
\begin{table}[h]
    \footnotesize
    \setlength{\tabcolsep}{3.5pt}
    \centering

    \begin{tabular}{l l ccc ccc}
        \toprule
        \multirow{2}{*}{\textbf{Method}} & \multicolumn{3}{c}{\textbf{ACID}} & \multicolumn{3}{c}{\textbf{DTU}} \\
        \cmidrule(lr){2-4} \cmidrule(lr){5-7}  
        &  PSNR$\uparrow$ & SSIM$\uparrow$ & LPIPS$\downarrow$ & PSNR$\uparrow$ & SSIM$\uparrow$ & LPIPS$\downarrow$ \\
        \midrule
        pixelSplat & 25.477 & 0.770 & 0.207 & 15.067 & 0.539 & 0.341  \\
         MVSplat & 25.525 & 0.773 & 0.199 & 14.542 & 0.537 & 0.324 \\
         NoPoSplat$^\ast$ & 25.765 & 0.776 & 0.199 & \underline{17.899} & \underline{0.629} & 0.279 \\
         SelfSplat & 22.204 & 0.686 & 0.316 & 13.249 & 0.434 & 0.441\\
         PF3plat & 20.726 & 0.610 & 0.308 & 12.972 & 0.407 & 0.464\\
         \textbf{SPFSplat} & \underline{25.965}	& \underline{0.781}	& \underline{0.190} &	16.550	& 0.579	 & \underline{0.270} \\
         \textbf{SPFSplat$^\ast$} & \textbf{26.697} & \textbf{0.806} & \textbf{0.181} & \textbf{18.297} & \textbf{0.660} & \textbf{0.255} \\
        \bottomrule
    \end{tabular}
    \vspace{-7pt}
    \caption{Cross-dataset generalization. All methods are trained on RE10K and evaluated in a zero-shot setting on ACID and DTU. Our method demonstrates superior generalization compared to SOTA approaches, even outperforming NoPoSplat’s ACID-trained model (PSNR: 25.961) as reported in Tab.~\ref{tab:acid_results}.}
\label{tab:cross-dataset}
\vspace{-20pt}
\end{table}

\noindent\textbf{Cross-Dataset Generalization.} 
To assess zero-shot generalization, we train exclusively on RE10K (indoor scenes) and evaluate on ACID (outdoor scenes) and DTU (object-centric scenes). The results in Tab.~\ref{tab:cross-dataset} and Fig.~\ref{fig:cross_dataset} demonstrate that our approach outperforms all SOTA methods. 
With no ground-truth poses used during training, our model learns 
to align the Gaussians with the predicted poses, enabling strong generalization to out-of-distribution scenes.

% \begin{figure*}[!ht]
%     \centering
%     \includegraphics[width=1.0\textwidth]{figs/3d_vis.pdf}
%     \vspace{-8pt}
%     \caption{Comparisons of 3D Gaussians and rendered results. The red and green indicate input and target camera poses, and the rendered image and depths are shown on the right side. Our method produces higher-quality 3D Gaussians and better rendering over baselines.}
%     \label{fig:3d_vis}
%     \vspace{-5pt}
% \end{figure*}

% \begin{figure}[!ht]
%     \centering
%     \includegraphics[width=\linewidth]{ICCV2025-Author-Kit-Feb/figs/3d_vis_v3.pdf}
%     \vspace{-20pt}
%     \caption{Comparisons of 3D Gaussians and rendered results. The red and green indicate input and target camera poses, and the rendered image and depths are shown on the right side. Our method produces higher-quality 3D Gaussians and better rendering over NoPoSplat.}
%     \label{fig:3d_vis}
%     \vspace{-15pt}
% \end{figure}

% \begin{figure*}[!ht]
%     \centering
%     \includegraphics[width=\linewidth]{ICCV2025-Author-Kit-Feb/figs/3d_vis_kitchen.pdf}
%     \vspace{-20pt}
%     \caption{Comparisons of 3D Gaussians and rendered results. The red and green indicate input and target camera poses, and the rendered image and depths are shown on the right side. Our method produces higher-quality 3D Gaussians and better rendering over baselines.}
%     \label{fig:3d_vis}
%     \vspace{-15pt}
% \end{figure*}

\begin{figure*}[!ht]
    \centering
    \includegraphics[width=\linewidth]{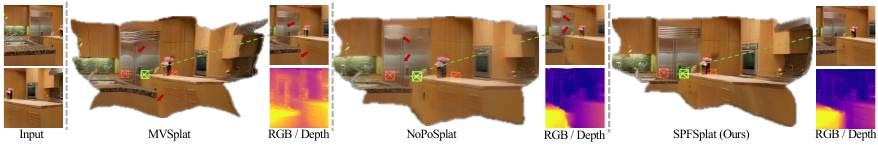}
    \vspace{-20pt}
    \caption{Comparison of 3D Gaussians and rendered results. Input and target camera poses are shown in red and green, respectively. Rendered images and depth maps are displayed on the right. Our method produces higher-quality 3D Gaussians and achieves superior rendering compared to baseline methods. Some regions with distorted or incorrect geometry are highlighted with red arrows.}
    \label{fig:3d_vis}
    \vspace{-15pt}
\end{figure*}

\noindent\textbf{Geometry Reconstruction.}
As shown in Fig.~\ref{fig:3d_vis}, our SPFSplat generates higher-quality 3D Gaussian primitives than baseline methods, despite being trained without ground-truth poses. The reconstructed structures are sharper and more accurate, indicating better Gaussian alignment across views. This improvement mainly stems from jointly optimizing Gaussians and camera poses, which encourages a stronger understanding of scene geometry.

\subsection{Ablation Analysis}
\label{sec:experiments_ablation}
\textbf{Ablation on Different Components.}
We conduct an ablation study to assess the contribution of each component in our method, as shown in Tab.~\ref{tab:ablation}.
As seen from (a) to (b), removing intrinsic embeddings from the backbone slightly reduces performance due to scale ambiguity in both 3D Gaussian learning and pose estimation. However, even without intrinsic embeddings, our method still outperforms NoPoSplat with intrinsic embeddings (PSNR: 25.033).
Comparing (a) and (c), removing the reprojection loss while retaining only the image rendering loss on target images significantly degrades both NVS and pose estimation performance, highlighting the importance of geometric constraints between 3D points and camera poses for accurate reconstruction.
% From (a) to (d), we replace the reprojection loss with an image rendering loss on context images results in a slight decline in NVS performance and a significant drop in pose estimation accuracy.
 \vspace{-5pt}
\begin{table}[h]
\footnotesize
\setlength{\tabcolsep}{3.2pt}
    \centering
    \begin{tabular}{l ccc ccc}
        \toprule
        \multirow{2}{*}{\textbf{Method}} & \multicolumn{3}{c}{\textbf{NVS$^\ast$}} & \multicolumn{3}{c}{\textbf{Pose}} \\
        \cmidrule(lr){2-4} \cmidrule(lr){5-7}  
        & PSNR$\uparrow$ & SSIM$\uparrow$ & LPIPS$\downarrow$ &  5$^\circ$ $\uparrow$ & 10$^\circ$ $\uparrow$ & 20$^\circ$ $\uparrow$  \\
        \midrule
        (a) SPFSplat (Ours) & \underline{25.845} & \underline{0.852} & \underline{0.152} & \underline{0.617} & \underline{0.755}	& \underline{0.845} \\
        (b) w/o intrin. emb.  &  25.519 & 0.844 & 0.156 & {0.562} & {0.717} & {0.823} \\
        (c) w/o reproj. loss  &  21.914 & 0.742 & 0.251 & 0.028 & 0.102 & 0.263 \\
        % (d) w/ context image loss &  &  &  & 0.547 & 0.701 & 0.807 \\
        (d) w/ gt pose loss & \textbf{25.910} & \textbf{0.860} & \textbf{0.150} & \textbf{0.691} & \textbf{0.810} & \textbf{0.885}\\
        \bottomrule
    \end{tabular}
    \vspace{-8pt}
\caption{Component ablations on RE10K. $\ast$ indicates the use of EPA strategy. See the appendix for NVS results without EPA.}
\label{tab:ablation}
\vspace{-10pt}
\end{table}

\noindent\textbf{Ablation on Ground-truth Poses.}
% \label{sec:ablation_on_gt_pose}
To evaluate our method’s ability to reconstruct geometry without pose priors, we introduce a pose loss during training that minimizes the difference between predicted and ground-truth poses (details provided in the appendix). This supervision is used only during training, keeping the method pose-free at inference.
As shown in Tab.~\ref{tab:ablation} (a) to (d), pose supervision improves pose accuracy but has only a marginal effect on NVS performance, highlighting our model’s strong capacity to reconstruct geometry without explicit pose supervision. It also suggests that NVS quality depends on factors beyond pose accuracy: challenges such as occlusion, texture-less regions, and extreme viewpoint changes may require generative abilities or explicit 3D supervision.

\noindent\textbf{Scale to More Training Data.} Since our approach does not require ground-truth poses for training, it can scale efficiently to larger training datasets with minimal additional cost. To assess the impact of training data size, we train on a combination of RE10K and DL3DV.  As shown in Tab.~\ref{tab:data_size}, increasing the amount of training data improves performance on both RE10K and ACID. This improvement is likely due to the increased diversity of camera motions in DL3DV, which enhances the model’s ability to generalize across different viewing conditions.
% \begin{table}[h]
% \footnotesize
%     \centering
%     \setlength{\tabcolsep}{5pt}
    
%     \vspace{-7pt}
%     % \renewcommand{\arraystretch}{1.2} % Adjust row spacing
%      \begin{tabular}{l ccc ccc}
%         \toprule
%         \multirow{2}{*}{\textbf{Training data}} & \multicolumn{3}{c}{\textbf{Pose Head}} & \multicolumn{3}{c}{\textbf{PnP Pose}} \\
%         \cmidrule(lr){2-4} \cmidrule(lr){5-7}  
%         & 5$^\circ$ $\uparrow$ & 10$^\circ$ $\uparrow$ & 20$^\circ$ $\uparrow$ & 5$^\circ$ $\uparrow$ & 10$^\circ$ $\uparrow$ & 20$^\circ$ $\uparrow$  \\

%         \midrule
%         Re10k & 0.600 & 0.742 & 0.838 & 0.609 & 0.746 & 0.841 \\
%         Re10k + DL3DV  &  \textbf{0.614} & \textbf{0.754} & \textbf{0.846} & \textbf{0.621} & \textbf{0.757} & \textbf{0.848} \\
%         \bottomrule
%     \end{tabular}
%     \vspace{-7pt}
% \caption{Ablation on training data size.}
% \vspace{-7pt}
% \label{tab:data_size}
% \end{table}

\begin{table}[h]
\footnotesize
    \centering
    \setlength{\tabcolsep}{5pt}
    
    \vspace{-7pt}
     \begin{tabular}{l ccc ccc}
        \toprule
        \multirow{2}{*}{\textbf{Training data}} & \multicolumn{3}{c}{\textbf{RE10K}} & \multicolumn{3}{c}{\textbf{ACID}} \\
        \cmidrule(lr){2-4} \cmidrule(lr){5-7}  
        & 5$^\circ$ $\uparrow$ & 10$^\circ$ $\uparrow$ & 20$^\circ$ $\uparrow$ & 5$^\circ$ $\uparrow$ & 10$^\circ$ $\uparrow$ & 20$^\circ$ $\uparrow$  \\

        \midrule
        Re10K & 0.617 & 0.755	& 0.845 & 0.364 & 0.520	& 0.662 \\
        Re10K + DL3DV  &  \textbf{0.635} &	\textbf{0.768} & \textbf{0.852} &	\textbf{0.395} & \textbf{0.544} & \textbf{0.680} \\
        \bottomrule
    \end{tabular}
    \vspace{-7pt}
\caption{Ablation on training data size.}
\vspace{-7pt}
\label{tab:data_size}
\end{table}

% \vspace{-4pt}

% \begin{table}[h]
% \footnotesize
%     \centering
%     \caption{Ablation on ground-truth poses.}
%     % \renewcommand{\arraystretch}{1.2} % Adjust row spacing
%     \begin{tabular}{lccc}
%         \toprule
%         & \textbf{PSNR$\uparrow$} & \textbf{SSIM$\uparrow$} & \textbf{LPIPS$\downarrow$} \\
%         \midrule
%         Ours & \textbf{25.316} & \textbf{0.843} & \textbf{0.155} \\
%         Ours + GT poses &   &  &  \\
%         \bottomrule
%     \end{tabular}
% \label{tab:gt_pose}
% \end{table}

% \textbf{Model Efficiency.}
% \begin{table}[h]
% \footnotesize
%  \setlength{\tabcolsep}{2.8pt}
%     \centering
%     \caption{Time.}
%     % \renewcommand{\arraystretch}{1.2} % Adjust row spacing
%     \begin{tabular}{cccccc}
%         \toprule
%         \textbf{pixelSplat} & \textbf{MVSplat} & \textbf{NopoSplat} & \textbf{SelfSplat}
%         & \textbf{Pf3plat} & \textbf{Ours}\\
%         \midrule
%         0.152 & 0.059 & 0.042 &  0.101
%         & 1.171 & 0.044\\
%         \bottomrule
%     \end{tabular}
% \label{tab:gt_pose}
% \end{table}

% \vspace{-8pt}
\noindent\textbf{Extension to Multiple Views.}
Our method extends naturally to multiple input views. As shown in Table~\ref{tab:multi_view_for_nvs}, quantitative results demonstrate that NVS performance improves consistently with an increasing number of context views.

% Performance improves significantly when increasing context views from two to three, with minimal gains from additional views, indicating that our method can effectively leverage information from three views to reconstruct the scene.

 \vspace{-7pt}
\begin{table}[h]
\footnotesize
    \centering
    \begin{tabular}{lccc}
        \toprule
        Num of Views & PSNR$\uparrow$ & SSIM$\uparrow$ & LPIPS$\downarrow$ \\
        \midrule
        2 views & 25.403 & 0.845 & 0.154 \\
        % 2 views & 25.631 & 0.850 & 0.151 \\
        3 views  &  26.724 & 0.871 & 0.128 \\
        % 4 views  &  26.831 & 0.874 & 0.125 \\
        5 views  & {26.891} & {0.875} & {0.122} \\
        10 views  &  \textbf{27.159} &  \textbf{0.880} & \textbf{0.115} \\
        \bottomrule
    \end{tabular}
     \vspace{-5pt}
     \caption{Novel view synthesis with varying input view numbers.}
     \vspace{-10pt}
\label{tab:multi_view_for_nvs}
\end{table}
 \vspace{-6pt}

% \begin{table*}[ht]
% \centering
% \caption{Pose estimation performance in AUC with various thresholds on RE10k and ACID datasets.}

% \begin{tabular}{lccccccccccc}
% \toprule
% \textbf{Method} & \multicolumn{3}{c}{\textbf{Pose Head}} & \multicolumn{3}{c}{\textbf{PnP Pose}}  \\
% \cmidrule(lr){2-4} \cmidrule(lr){5-7}  

% &  5$^\circ$ $\uparrow$ & 10$^\circ$ $\uparrow$ & 20$^\circ$ $\uparrow$ &  5$^\circ$ $\uparrow$ &  10$^\circ$ $\uparrow$ &  20$^\circ$ $\uparrow$  \\
% \midrule
% % CoPoNeRF & 0.161 & 0.362 & 0.575 & 0.078 & 0.216 & 0.398 \\

% \textbf{Ours} & 0.470 & 0.661 & 0.794 & 0.314 & 0.488 & 0.640 \\
% \textbf{Ours} & \textbf{0.611} & \textbf{0.748} & \textbf{0.842} & \textbf{0.356} & \textbf{0.516} & \textbf{0.656} \\
% % \midrule
% % \textbf{Ours (RE10k) + gt} & 0.598 & 0.750 & 0.848 & 0.386 & 0.546 & 0.676 \\
% \bottomrule
% \end{tabular}

% \end{table*}
\section{Conclusion}
\label{sec:conclusion}
This paper introduces SPFSplat, an efficient self-supervised pose-free framework designed for sparse-view 3D reconstruction. It employs a shared backbone to simultaneously predict 3D Gaussian representations and camera poses in a canonical space given unposed input views. A reprojection loss is also incorporated with the conventional rendering loss to enhance geometric alignment. Experimental evaluations highlight the superior performance of SPFSplat in novel view synthesis, relative pose estimation, and zero-shot generalization. Notably, the independence from ground-truth pose annotations underscores its potential for scalable training on large-scale real-world data.

\section*{A More Implementation Details}
\label{sec:appendix_details}

\textbf{More Training Details.}
We use a batch size of 12, with each batch containing one training scene consisting of input views and target views. As training progresses, the frame distance between input views gradually increases. The initial learning rate is set to $1 \times 10^{-5}$ for the backbone and $1 \times 10^{-4}$ for other parameters.

\begin{figure}[h]
    \centering
    \includegraphics[width=\linewidth]{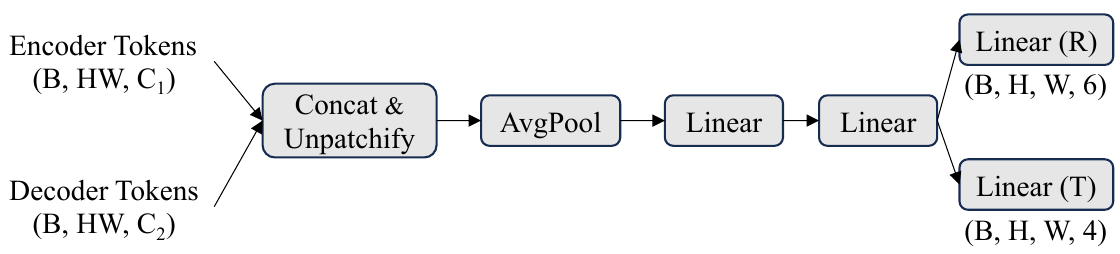}
    \caption{The structure of the proposed pose head.}
    \label{fig:pose_head}
    \vspace{-8pt}
\end{figure}

\noindent\textbf{More Details of Pose Head.}
The pose head structure is shown in Fig.~\ref{fig:pose_head}. During training, the linear layer for rotation is initialized with zero weights, and the 6D bias is set to $(1,0,0,0,1,0)$ to approximate the identity matrix, ensuring that the initial pose for each view has a shared field of view for stable convergence. Camera normalization sets the pose of the first view to the identity matrix. 
% To address scale ambiguity, we normalize the baseline of predicted input view poses to unit length.

\noindent\textbf{More Details of Splatt3R Results.}
In Tab.~\ref{tab:rek_results} and Tab.~\ref{tab:acid_results} of the main paper, we retrain Splatt3R since its original implementation does not provide results on RE10K and ACID. We remove the loss mask, as it depends on ground-truth depth, which is unavailable for RE10K.  Additionally, the original implementation employs an offset head to adjust the 3D points predicted by the frozen MASt3R. However, we find that this approach inefficient for aligning with the scale of ground-truth intrinsics. Instead, we directly finetune the 3D point head while keeping the backbone frozen.
Training is conducted using ground-truth poses.
% with the baseline normalized to one.

\noindent\textbf{More Details of Baselines in Pose Estimation.}
In Tab.\ref{tab:pose_estimation}, all experiments are conducted using 256$\times$256 input images. For SuperPoint + SuperGlue, feature matches are used to estimate Essential Matrices and compute relative poses.
For DUSt3R and MASt3R, we estimate camera intrinsics from the 3D points of the first view and compute relative poses using the PnP algorithm~\cite{hartley2003multiple}.
Since SelfSplat defines the target image as the reference frame, we evaluate relative poses from the target to the context image, a relatively easier task than predicting relative poses between two context views.
NoPoSplat estimates poses in two stages: it first initializes the relative pose between input views using PnP~\cite{hartley2003multiple} with RANSAC~\cite{fischler1981ransac}, leveraging predicted Gaussian centers. Then, with the Gaussian parameters fixed, it refines the pose by minimizing photometric losses combined with an SSIM loss. This second-stage optimization integrates 3D Gaussian splatting into the loop, making it computationally expensive and less suitable for real-time applications.
For a fair comparison with other splatting-based methods, we report NoPoSplat’s accuracy based on the first-stage initialization only.

\noindent\textbf{More Details of Ablation on Ground-truth Poses.}
To evaluate our method’s ability to reconstruct geometry without pose supervision, as shown in Tab.~\ref{tab:ablation}, we incorporate camera poses to supervise our pose head. Our pose loss is a combination of geodesic loss~\cite{salehi2018geo} for rotation and $L_2$ distance loss for translation. Specifically, they are defined in Eq.~\ref{eq:rot_loss} and Eq.~\ref{eq:trans_loss}. We set the weight for rotation loss to 0.1, the weight for translation loss to 0.01.

\begin{align}
    &\mathcal{L}_{\text{rot}} = \arccos \left( \frac{\operatorname{trace}(\boldsymbol{\hat{R}}^T \boldsymbol{R}) - 1}{2} \right) \label{eq:rot_loss} \\
    &\mathcal{L}_{\text{trans}}(\boldsymbol{\hat{T}}, \boldsymbol{T}) = \| \boldsymbol{\hat{T}} - \boldsymbol{T} \|_2^2 \label{eq:trans_loss}
\end{align}

% \noindent\textbf{Evaluation-time Pose Alignment}
% NoPoSplat~\cite{ye2025noposplat} adopts evaluation-time pose alignment (EPA) strategy, optimizing the target camera pose during evaluation. Specifically, after reconstructing 3D Gaussians, the Gaussians are frozen, they set the initial target pose using the PnP algorithm (Hartley & Zisserman, 2003) with RANSAC (Fischler & Bolles, 1981), given the Gaussian centers of the output Gaussians in world coordinates. and only the target pose is optimized to ensure that the rendered image most closely matches the ground truth. 
% For EPA evaluation in our paper, we set the initial target pose as the predicted pose of the pose head.

\section*{B More Experimental Analysis}
\label{sec:appendix_experiments}

% \noindent\textbf{Evaluation on More Datasets.}
% Tab.~\ref{tab:dl3dv_scannetpp} shows our method outperforms all SOTA methods on the DL3DV and ScanNet++ datasets.
% \input{table_sup/dl3dv_scannetpp}

\begin{figure}[ht]
    \centering
    \includegraphics[width=\linewidth]{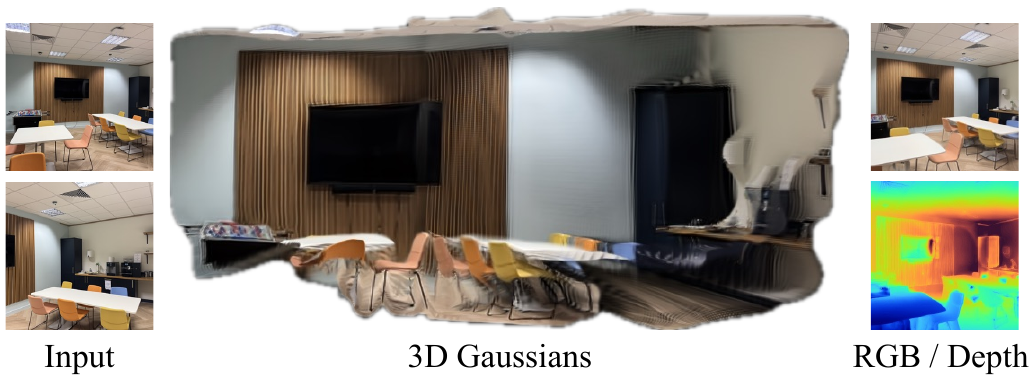}
    \caption{3D Gaussians and rendered RGB and depth results on mobile phone photos.}
    \label{fig:phone_vis}
    \vspace{-10pt}
\end{figure}

\noindent\textbf{Evaluation on In-the-Wild Data.}
We evaluate our model on mobile phone photos using the version trained without intrinsic embeddings (as in the ablation study). Given two input images, we estimate the focal length from the output Gaussian centers of the canonical view (the first image). This focal length is used to render novel views. The 3D geometry and rendered results in Fig.~\ref{fig:phone_vis} demonstrate our model’s strong out-of-domain generalization, even under large viewpoint changes.

\noindent\textbf{Evaluation on the Evaluation Set of pixelSplat.}
We adopt the evaluation set from NoPoSplat~\cite{ye2025noposplat} in our main paper, as it presents a greater challenge due to minimal overlap between most input pairs. Additionally, we report results using the evaluation sets from pixelSplat~\cite{charatan2024pixelsplat} and MVSplat~\cite{chen2024mvsplat}, as shown in Tab.~\ref{tab:pixelsplat_evaluation_set}.These results also demonstrates that our method consistently outperforms other SOTA methods.
\begin{table}[ht]
    \footnotesize
    \centering
    \setlength{\tabcolsep}{4pt}
    \begin{tabular}{l l ccc ccc}
        \toprule
        \multirow{2}{*}{\textbf{Method}} & \multicolumn{3}{c}{\textbf{RE10K}} & \multicolumn{3}{c}{\textbf{ACID}} \\
        \cmidrule(lr){2-4} \cmidrule(lr){5-7}  
        &  PSNR$\uparrow$ & SSIM$\uparrow$ & LPIPS$\downarrow$ & PSNR$\uparrow$ & SSIM$\uparrow$ & LPIPS$\downarrow$ \\
        \midrule
        pixelSplat & 26.090 & 0.863 & 0.136 & 28.270	& 0.843	& 0.146 \\
         MVSplat &  26.387 & 0.869 & 0.128 & 28.254 & 0.843 & 0.144 \\
        % & NopoSplat & 23.379 & 0.684 & 0.237 & 14.034 & 0.414 & 0.503 \\
         NopoSplat$^\ast$ & 26.820 & 0.880 & 0.125 & 27.952 & 0.837 & 0.150 \\
         Ours & \underline{27.311} & \underline{0.888} &	\underline{0.119} & \underline{28.407} &	\underline{0.845}	& \underline{0.142} \\
         Ours$^\ast$ & \textbf{27.696} &	\textbf{0.892} &	\textbf{0.117}	& \textbf{29.000} &	\textbf{0.862} &	\textbf{0.136} \\
         % Ours & \underline{27.469}  & \underline{0.890} & \underline{0.118} & \underline{28.360} & \underline{0.844} & \underline{0.144} \\
         % Ours + EPA & \textbf{27.777} & \textbf{0.893} & \textbf{0.116} & \textbf{28.910} & \textbf{0.859} & \textbf{0.138} \\
        \bottomrule
    \end{tabular}
     \caption{Novel view synthesis performance comparison on the evaluation sets of pixelSplat and MVSplat.}
     \vspace{-10pt}
\label{tab:pixelsplat_evaluation_set}
\end{table}

\noindent\textbf{Comparison on PnP Pose and Pose Head.}
Our method supports two strategies for pose estimation: direct regression via the pose head, and estimation via PnP~\cite{hartley2003multiple} with RANSAC~\cite{fischler1981ransac}, using the predicted 3D Gaussian centers. As shown in Tab.~\ref{tab:comparison_pnp_pred}, both approaches achieve comparable results in both in-domain and out-of-domain settings, indicating strong alignment between the estimated poses and the predicted Gaussian centers.
\begin{table}[ht]
    \footnotesize
    \centering
    \setlength{\tabcolsep}{2pt}
    \begin{tabular}{l l ccc ccc ccc}
        \toprule
        \multirow{2}{*}{\textbf{Task}} & \multirow{2}{*}{\textbf{Method}} & \multicolumn{3}{c}{\textbf{Rotation}} & \multicolumn{3}{c}{\textbf{Translation}}  \\
        \cmidrule(lr){3-5} \cmidrule(lr){6-8} 
        &&  5$^\circ$ $\uparrow$ & 10$^\circ$ $\uparrow$ & 20$^\circ$ $\uparrow$ &  5$^\circ$ $\uparrow$ &  10$^\circ$ $\uparrow$ &  20$^\circ$ \\
        \midrule
        % \rowcolor{gray!20} \multicolumn{10}{l}{\textit{Supervised Pose-Free}} \\ 
        \multirow{2}{*}{RE10K $\rightarrow$ RE10K } & PnP & 0.793 & 0.875 & 0.926 & 0.661	& 0.789	& 0.872  \\
         & Pose Head &  0.816 & 0.886 & 0.932 & 0.666 & 0.793 & 0.874 \\
         \midrule
         \multirow{2}{*}{RE10K $\rightarrow$ ACID } & PnP & 0.614 & 0.739 & 0.830 & 0.384	& 0.545	& 0.683  \\
         & Pose Head &  0.645 & 0.754 & 0.838 & 0.402 & 0.555 & 0.689 \\

        %  \midrule
        % \multirow{2}{*}{RE10K $\rightarrow$ RE10K } & PnP & 0.774 & 0.857 & 0.910 & 0.653	& 0.779	& 0.862  \\
        %  & Pose Head &  0.777 & 0.858 & 0.911 & 0.659 & 0.782 & 0.864 \\
        %  \midrule
        %  \multirow{2}{*}{RE10K $\rightarrow$ ACID } & PnP & 0.622 & 0.739 & 0.830 & 0.394	& 0.551	& 0.682  \\
        %  & Pose Head &  0.643 & 0.750 & 0.835 & 0.403 & 0.556 & 0.686 \\

        \bottomrule
    \end{tabular}
      \caption{Comparison between PnP poses estimated from Gaussian centers and poses predicted by the pose head.}
\label{tab:comparison_pnp_pred}
\end{table}

\begin{figure}[ht]
    \centering
    \includegraphics[width=0.48\textwidth]{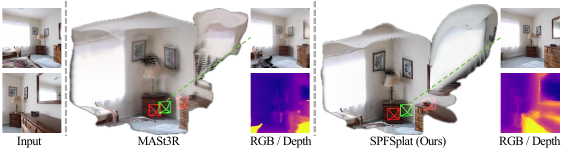}
    \caption{Comparison of 3D Gaussian and rendered results  between MASt3R and our method.}
    \label{fig:mast3r_3d_vis}
    \vspace{-8pt}
\end{figure}
% \begin{table*}[!htb]
% \footnotesize
% \centering
% % \renewcommand{\arraystretch}{1.2}
% % \captionsetup{justification=centering, singlelinecheck=false} % Caption spans full width
% \label{tab:re10k_nvs}
% \resizebox{\textwidth}{!}{ % Resize to fit within page width
% \begin{tabular}{lcccccccccccc}
% \toprule
% \textbf{Method} & \multicolumn{3}{c}{\textbf{Small}} & \multicolumn{3}{c}{\textbf{Medium}} & \multicolumn{3}{c}{\textbf{Large}} & \multicolumn{3}{c}{\textbf{Average}} \\
% & {PSNR $\uparrow$} & {SSIM $\uparrow$} & {LPIPS $\downarrow$} 
% & {PSNR $\uparrow$} & {SSIM $\uparrow$} & {LPIPS $\downarrow$} 
% & {PSNR $\uparrow$} & {SSIM $\uparrow$} & {LPIPS $\downarrow$} 
% & {PSNR $\uparrow$} & {SSIM $\uparrow$} & {LPIPS $\downarrow$} \\ 
% \midrule

% MASt3R & 16.305 & 0.516 & 0.451 
% & 18.106 & 0.561 & 0.377 
% & 17.975 & 0.524 & 0.402
% & 17.617 & 0.539 & 0.403\\

% \textbf{Ours} & \textbf{22.867} & \textbf{0.789} & \textbf{0.202} 
% & \textbf{25.258} & \textbf{0.846} & \textbf{0.154}	
% & \textbf{27.815} & \textbf{0.892} & \textbf{0.114} 
% & \textbf{25.403} & \textbf{0.845} & \textbf{0.154}\\

% \bottomrule
% \end{tabular}
% }
% \caption{Novel view synthesis performance comparison between MASt3R and our method on RE10K. }
% \label{tab:mast3r_rek_results}
% \end{table*}

\begin{table}[h]
\footnotesize
    \centering

    \begin{tabular}{lccc}
        \toprule
        Method & PSNR$\uparrow$ & SSIM$\uparrow$ & LPIPS$\downarrow$ \\
        \midrule
        MASt3R &  17.617 & 0.539 & 0.403 \\
        \textbf{SPFSplat (Ours)} & \textbf{25.484} & \textbf{0.847} & \textbf{0.153} \\
        \bottomrule
    \end{tabular}
\caption{Novel view synthesis performance comparison between MASt3R and our SPFSplat on RE10K. }
\label{tab:mast3r_rek_results}
\end{table}
\noindent\textbf{Comparison with MASt3R.}
As our method is initialized from MASt3R weights, to compare with the original MASt3R,
we adapt MASt3R for novel view synthesis by freezing its weights, setting its output 3D points as Gaussian centers, and adding a DPT head to predict other Gaussian parameters (as in our approach). To address scale inconsistency, we estimate the focal length from the 3D points and use it for training instead of the ground-truth focal length. The model is trained with ground-truth poses. As shown in Tab.~\ref{tab:mast3r_rek_results} and Fig.~\ref{fig:mast3r_3d_vis}, although our model is initialized from MASt3R and trained without pose supervision, it significantly outperforms MASt3R, achieving more accurate geometric structures and visual details.

\begin{table}[h]
\footnotesize
    \centering
   \setlength{\tabcolsep}{4.5pt}
    \begin{tabular}{lccc}
        \toprule
        Initialization & PSNR$\uparrow$ & SSIM$\uparrow$ & LPIPS$\downarrow$ \\
        \midrule
        Random & 21.200 & 0.690 & 0.250 \\
        DUSt3R &  25.280   &  0.841  &  0.156 \\
        \textbf{MASt3R} & \textbf{25.484} & \textbf{0.847} & \textbf{0.153} \\
        \bottomrule
    \end{tabular}
     % \vspace{-10pt}
 \caption{Comparison of different initialization strategies}
 \vspace{-10pt}
\label{tab:initialization}
\end{table}

% \begin{table}[h]
% \footnotesize
%     \centering
%    \setlength{\tabcolsep}{4.5pt}
%     % \renewcommand{\arraystretch}{1.2} % Adjust row spacing
%     \begin{tabular}{l ccc ccc}
%         \toprule
%         \multirow{2}{*}{\textbf{Initialization}} & \multicolumn{3}{c}{\textbf{NVS}} & \multicolumn{3}{c}{\textbf{Pose}} \\
%         \cmidrule(lr){2-4} \cmidrule(lr){5-7}  
%         & PSNR$\uparrow$ & SSIM$\uparrow$ & LPIPS$\downarrow$ &  5$^\circ$ $\uparrow$ & 10$^\circ$ $\uparrow$ & 20$^\circ$ $\uparrow$  \\
%         \midrule
%         Random & \\
%         DUSt3R &  25.280   &  0.841  &  0.156 & 0.602 & 0.750 & 0.846 \\
%         \textbf{MASt3R} & \textbf{25.484} & \textbf{0.847} & \textbf{0.153} & \textbf{0.617} & \textbf{0.755}	& \textbf{0.845}\\
%         \bottomrule
%     \end{tabular}
%  \caption{Comparisons of different initializations.}
%  \vspace{-8pt}
% \label{tab:initialization}
% \end{table}

\noindent\textbf{Initialization.}
In our main paper, we initialize the backbone with MASt3R weights. Here, we further analyze the influence of different backbone initialization strategies. As shown in Tab.~\ref{tab:initialization}, MASt3R’s pretrained weights achieve slightly better NVS performance compared to DUSt3R. This improvement can be attributed to MASt3R’s training on feature-matching tasks, which produces stronger local feature representations that improve both pose estimation accuracy and the quality of reconstructed 3D Gaussians. For random initialization, we adopt a warm-up phase by incorporating a point cloud distillation loss from the DUSt3R model during the first 10,000 steps. This additional supervision is crucial, as training with only photometric loss, especially without ground-truth geometric supervision, makes it difficult for the network to learn to predict Gaussians in the canonical space.
Since our model is trained without ground-truth poses, proper initialization significantly improves pose estimation quality. Although random initialization results in a noticeable performance drop, the results still demonstrate the model’s capability to reconstruct Gaussians without known poses.
% The initialization from MASt3R is crucial in the absence of geometric priors, as MASt3R excels in 3D pointmap prediction and feature matching, providing a superior initialization for both the Gaussians and the pose head. We also experimented with DUSt3R as pretrained weights but found it much inferior to MASt3R, as shown in Tab.~\ref{tab:initialization}. 

\begin{table}[h]
\footnotesize
\setlength{\tabcolsep}{3pt}
    \centering
    \begin{tabular}{l ccc ccc}
        \toprule
        \multirow{2}{*}{\textbf{Method}} & \multicolumn{3}{c}{\textbf{NVS}} & \multicolumn{3}{c}{\textbf{Pose}} \\
        \cmidrule(lr){2-4} \cmidrule(lr){5-7}  
        & PSNR$\uparrow$ & SSIM$\uparrow$ & LPIPS$\downarrow$ &  5$^\circ$ $\uparrow$ & 10$^\circ$ $\uparrow$ & 20$^\circ$ $\uparrow$  \\
        \midrule
        (a) Ours & \textbf{25.484} & \textbf{0.847} & \underline{0.153} & \underline{0.617} & \underline{0.755}	& \underline{0.845}\\
        % (a) Ours & \textbf{25.637} & \textbf{0.850} & \textbf{0.151} \\
        % (b) w/o intrin. emb.  &  24.439 & 0.817 & 0.168 \\
        (b) w/o intrin. emb.  &  24.864 & 0.829 & 0.161 & {0.562} & {0.717} & {0.823} \\
        (c) w/o reproj. loss  &  19.836 & 0.644 & 0.289  & 0.028 & 0.102 & 0.263 \\
       
        (d) w/ gt pose loss & 25.239 & \underline{0.842} & {0.157} & \textbf{0.691} & \textbf{0.810} & \textbf{0.885}\\
        (e) w/o $L_2$ loss  &   24.310 & 0.837  & \textbf{0.150}   & 0.602 & 0.740 & 0.833 \\
        (f) w/o LPIPS loss  &  \underline{25.336}    & 0.832 & 0.210  & 0.559 & 0.709 & 0.812  \\
        \bottomrule
    \end{tabular}
   \caption{Component ablations on RE10K. NVS are evaluated using predicted pose.}
\label{tab:ablation_without_epa}
\end{table}
\begin{figure}[ht]
    \centering
    \includegraphics[width=1.0\linewidth]{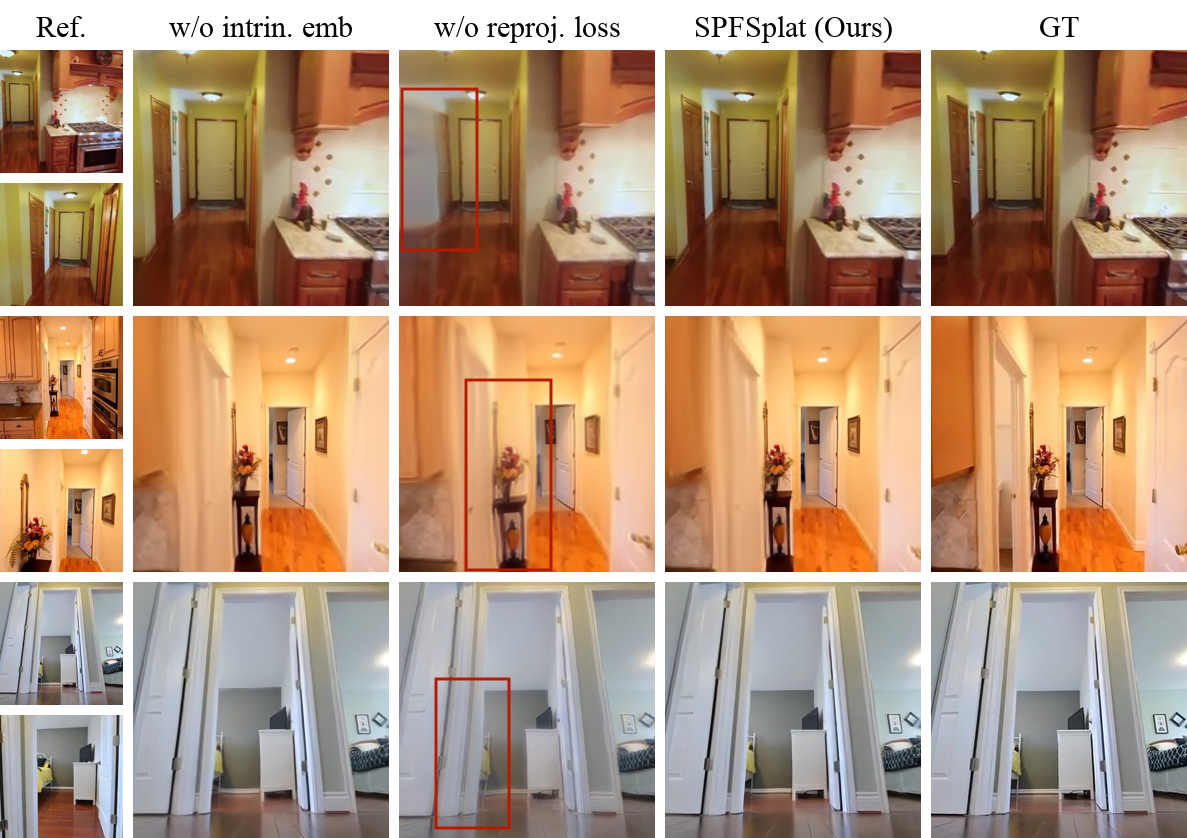}
    \caption{Ablation on the intrinsics embedding and reprojection loss.  Some failure regions are highlighted by red rectangles.}
    \label{fig:reproj_ablation}
\end{figure}

\begin{figure*}[ht]
    \centering
    %\vspace{-15pt}
    \includegraphics[width=1.0\textwidth]{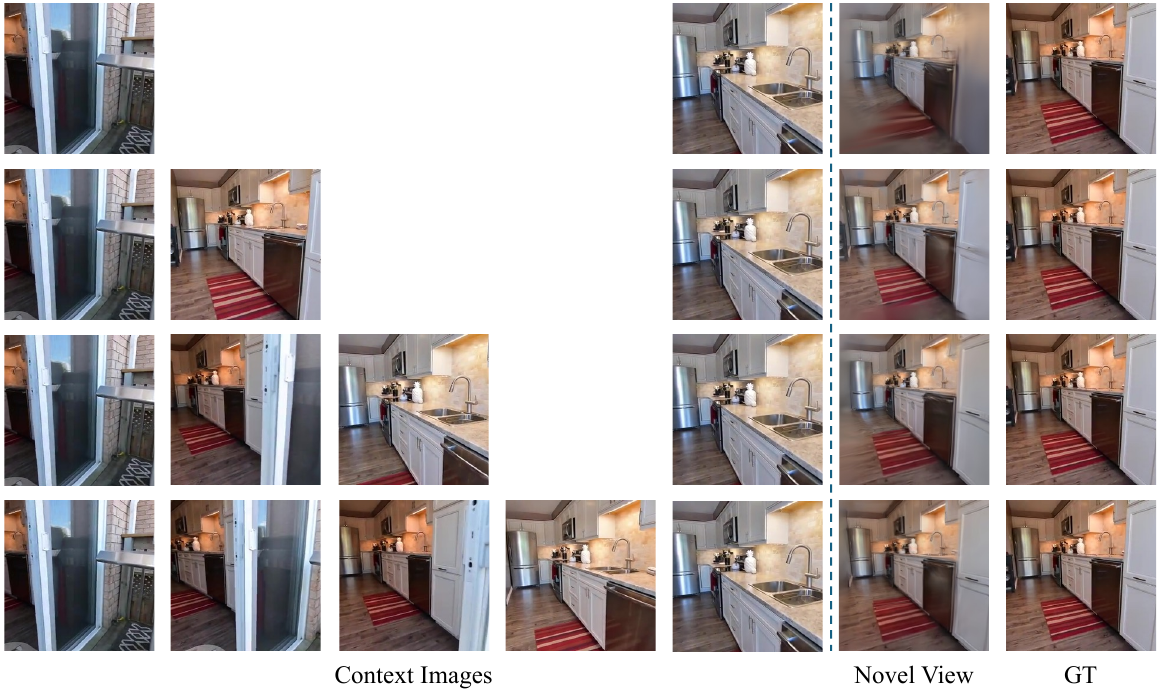}
    \caption{Qualitative comparison on different numbers of input views.}
    %\vspace{-15pt}
    \label{fig:multi_view}
\end{figure*}
\noindent\textbf{More Ablation Results.}
We demonstrate the ablation results on RE10K evaluated using predicted poses in Tab.~\ref{tab:ablation_without_epa}. It indicates that the intrinsics embedding and reprojection loss both contribute to better alignment between the poses and Gaussians. We also incorporate the ablation of L2 and LPIPS in (e) and (f), which demonstrates that both L2
and LPIPS losses positively impact NVS and pose estimation.
% \noindent\textbf{Ablation on Intrinsics Embedding and Reprojection Loss.}
Fig.~\ref{fig:reproj_ablation} presents the rendered results of our method without intrinsics embedding or reprojection loss. Removing intrinsics embedding leads to slightly blurrier outputs due to scale ambiguity. The absence of reprojection loss, however, results in severe blurring and rendering artifacts, as the lack of geometric constraints hinders proper alignment between poses and 3D points, causing reconstruction errors.

\noindent\textbf{Extension to Multiple Views.} Our method can be extended to multi-view input.
For a fair comparison, we fix the first and last views across all experiments while gradually increasing the number of intermediate views. As shown in Fig.~\ref{fig:multi_view}, more input views progressively enhance scene completeness and visual details, leading to improved rendering quality.

\begin{figure}[ht]
\vspace{-10pt}
    \centering
    \includegraphics[width=0.95\linewidth]{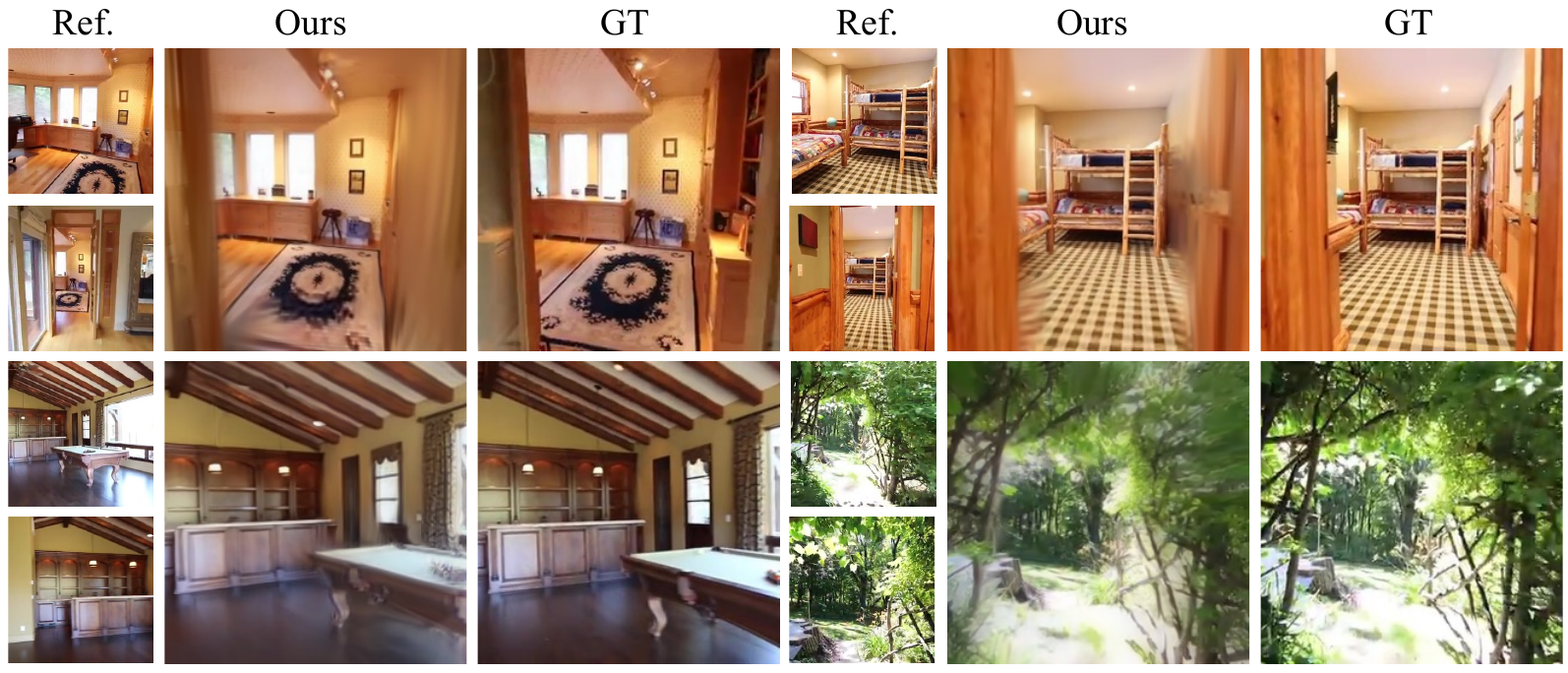}
    \vspace{-10pt}
    \caption{Some bad cases of our SPFSplat.}
    \label{fig:bad_cases}
    \vspace{-10pt}
\end{figure}

\noindent\textbf{{Failure Cases.}}
Fig.~\ref{fig:bad_cases} shows that our method may produce blurred outputs or artifacts in occluded or texture-less regions, or under extreme viewpoint changes. These issues may require generative abilities or explicit 3D supervision.

\section*{C More Visualizations}
\begin{figure*}[ht]
    \centering
    \includegraphics[width=0.98\textwidth]{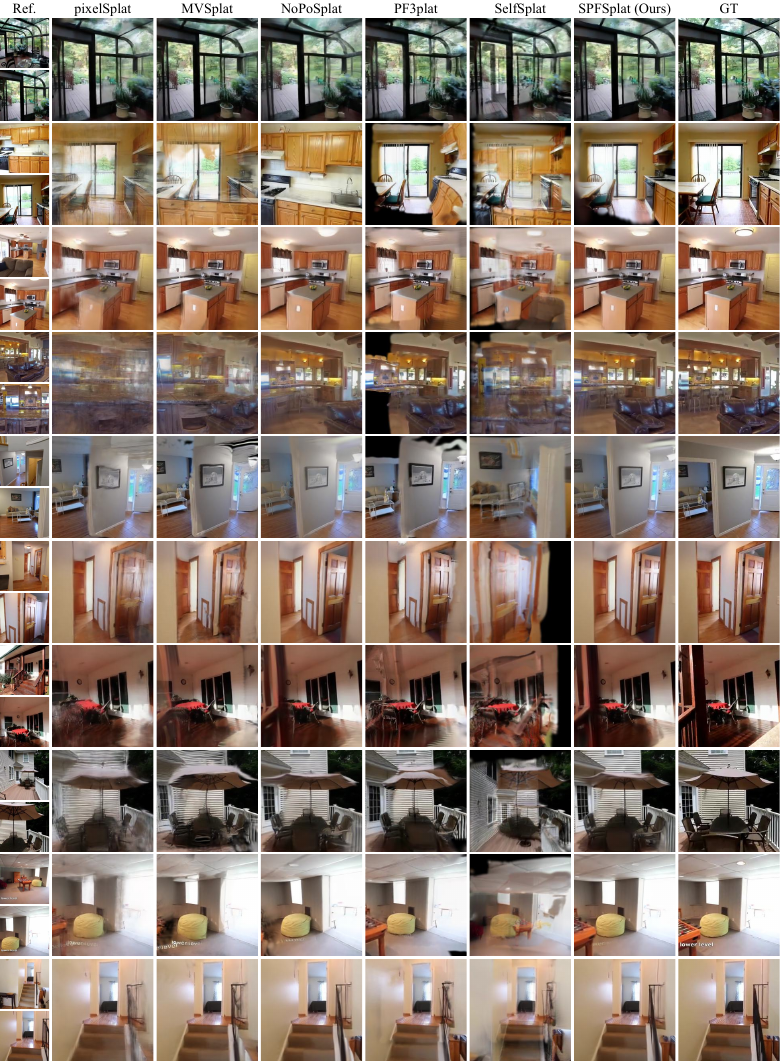}
    \vspace{-5pt}
    \caption{More qualitative comparisons on RE10K with small image overlap.}
    \label{fig:qualitative_comparison_re10k_small_sup}
\end{figure*}

\begin{figure*}[ht]
    \centering
    \includegraphics[width=0.98\textwidth]{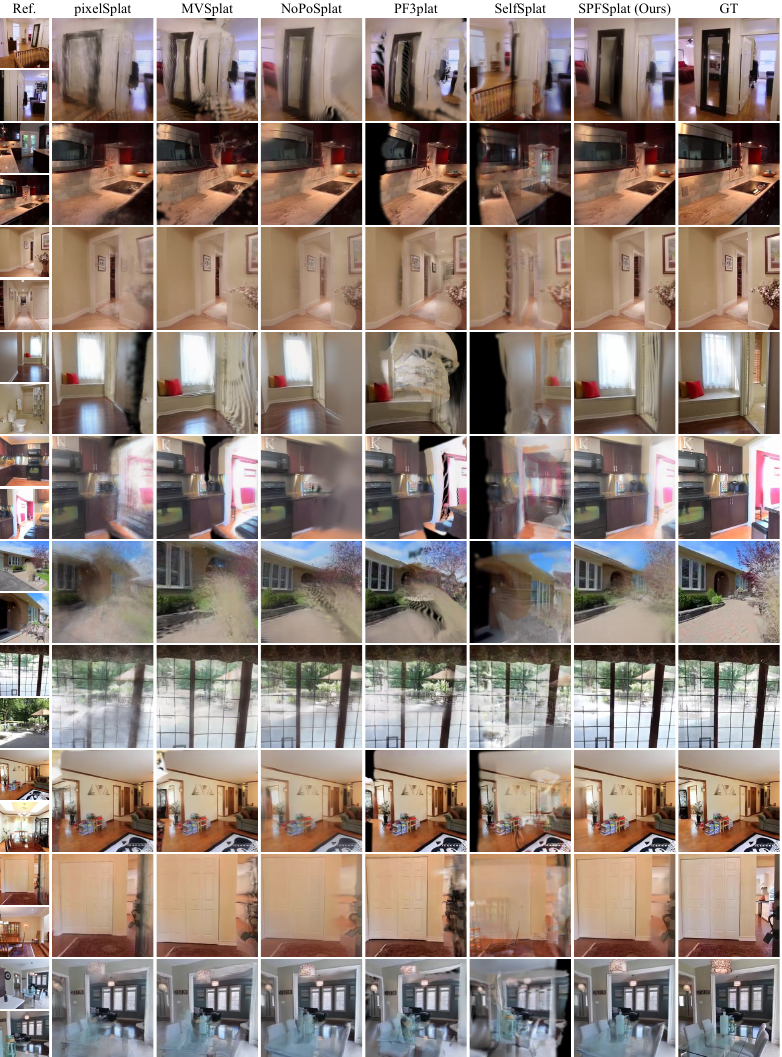}
    \vspace{-5pt}
    \caption{More qualitative comparisons on RE10K with medium image overlap.}
    \label{fig:qualitative_comparison_re10k_medium_sup}
\end{figure*}

\begin{figure*}[ht]
    \centering
    \includegraphics[width=0.98\textwidth]{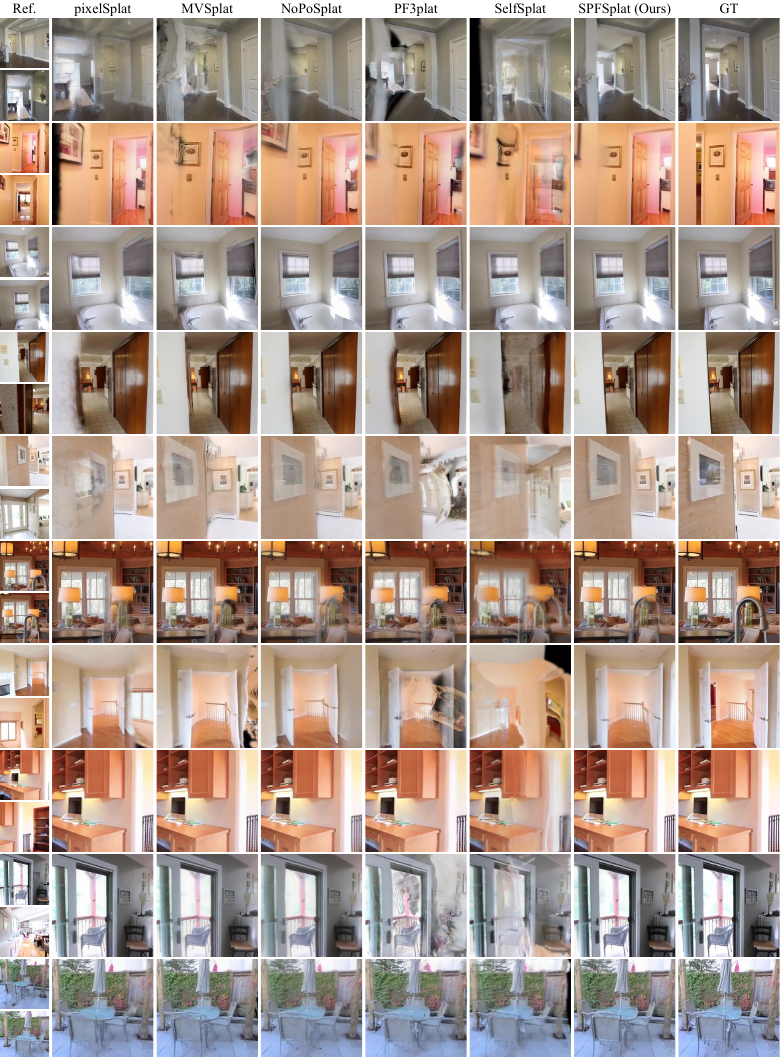}
    \vspace{-5pt}
    \caption{More qualitative comparisons on RE10K with large image overlap.}
    \label{fig:qualitative_comparison_re10k_large_sup}
\end{figure*}

\begin{figure*}[ht]
    \centering
    \includegraphics[width=0.98\textwidth]{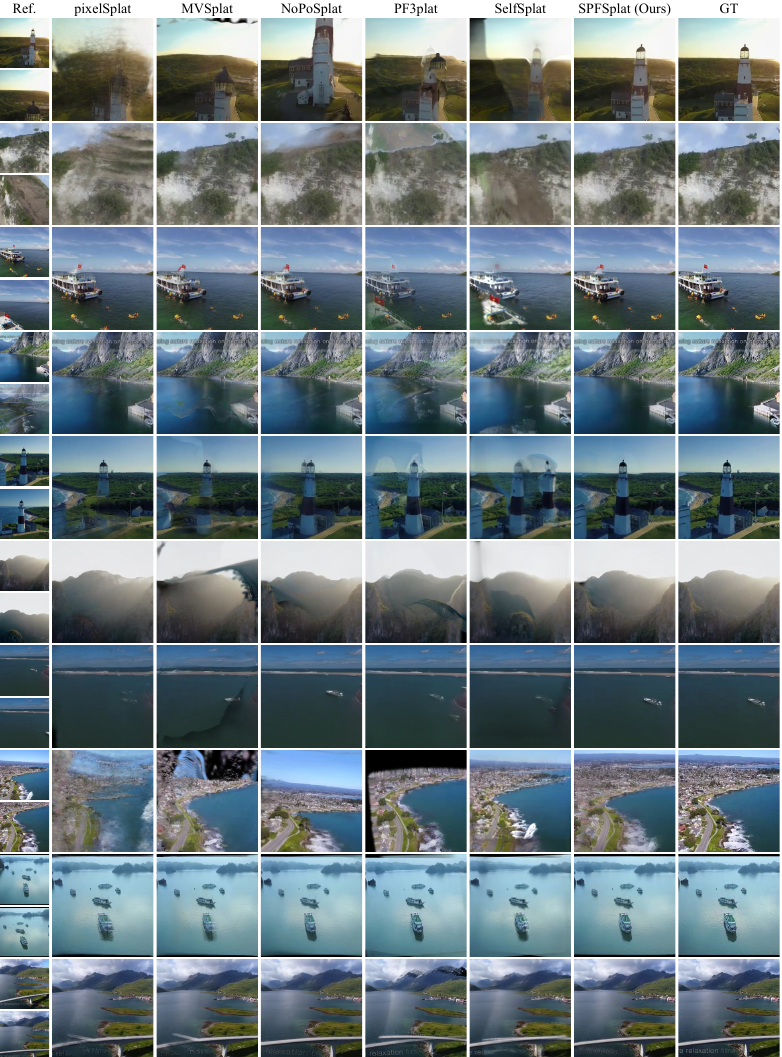}
    \vspace{-5pt}
    \caption{More qualitative comparisons on ACID.}
    \label{fig:qualitative_comparison_acid_sup}
\end{figure*}

We show more qualitative comparisons with baselines in Fig.~\ref{fig:qualitative_comparison_re10k_small_sup} to Fig.~\ref{fig:qualitative_comparison_acid_sup}.
Our method achieves stable and superior performance across different levels of image overlap and diverse datasets.

\section*{D Limitations}
Our method can be trained without ground-truth poses and easily scales to large datasets, therefore, future work could explore training on larger, more diverse datasets to improve pose estimation and generalization ability. Moreover, since our approach is not generative, it cannot reconstruct unseen areas with detailed textures. Generative models could be leveraged to mitigate this limitation.

% \clearpage
\clearpage
{
    \small
    \bibliographystyle{ieeenat_fullname}
    \bibliography{main}
}

\end{document}